\documentclass[journal,comsoc]{IEEEtran}
\usepackage{enumitem}
\usepackage{diagbox}
\usepackage[T1]{fontenc}
\usepackage{graphicx}
\usepackage{xspace}
\usepackage{algorithm}
\usepackage{algpseudocode}
\usepackage{multirow}
\usepackage{arydshln}
\usepackage{xcolor}
\usepackage{subfigure}
\usepackage{hyperref}
\usepackage{adjustbox}
\usepackage{dblfloatfix}
\graphicspath{{./figs/}}
\makeatletter
\DeclareRobustCommand\onedot{\futurelet\@let@token\@onedot}
\def\@onedot{\ifx\@let@token.\else.\null\fi\xspace}

\def\eg{\emph{e.g}\onedot}
\def\ie{\emph{i.e}\onedot}

\makeatother

\makeatletter
\newcommand*\bigcdot{\mathpalette\bigcdot@{.5}}
\newcommand*\bigcdot@[2]{\mathbin{\vcenter{\hbox{\scalebox{#2}{$\m@th#1\bullet$}}}}}
\makeatother

\newcommand{\Expect}{\operatorname{\mathbb{E}}}

\ifCLASSINFOpdf
\else
\fi

\usepackage{amsmath}
\usepackage[cmintegrals]{newtxmath}

\begin{document}

\title{ReDFeat: Recoupling Detection and Description for Multimodal Feature Learning}

\author{Yuxin Deng and Jiayi Ma
\thanks{This work was supported by the National Natural Science Foundation of China under Grant no. 61773295. \emph{(Corresponding author: Jiayi Ma.)}}
\IEEEcompsocitemizethanks{\IEEEcompsocthanksitem The authors are from the Electronic Information School, Wuhan University, Wuhan, 430072, China (e-mail: dyx\_acuo@whu.edu.cn, jyma2010@gmail.com).}
}


\maketitle

\begin{abstract}
Deep-learning-based local feature extraction algorithms that combine detection and description have made significant progress in visible image matching. However, the end-to-end training of such frameworks is notoriously unstable due to the lack of strong supervision of detection and the inappropriate coupling between detection and description. The problem is magnified in cross-modal scenarios, in which most methods heavily rely on the pre-training. In this paper, we recouple independent constraints of detection and description of multimodal feature learning with a mutual weighting strategy, in which the detected probabilities of robust features are forced to peak and repeat, while features with high detection scores are emphasized during optimization. Different from previous works, those weights are detached from back propagation so that the detected probability of indistinct features would not be directly suppressed and the training would be more stable. Moreover, we propose the Super Detector, a detector that possesses a large receptive field and is equipped with learnable non-maximum suppression layers, to fulfill the harsh terms of detection. Finally, we build a benchmark that contains cross visible, infrared, near-infrared and synthetic aperture radar image pairs for evaluating the performance of features in feature matching and image registration tasks. Extensive experiments demonstrate that features trained with the recoulped detection and description, named ReDFeat, surpass previous state-of-the-arts in the benchmark, while the model can be readily trained from scratch.
\end{abstract}


\section{Introduction}
\IEEEPARstart{F}{eature} detection and description are fundamental steps in many computer vision tasks, such as visual localization~\cite{zhang2021reference,zhang2021reference}, Structure-from-Motion (SfM)~\cite{schonberger2016structure}, and Simultaneous-Localization-and-Mapping (SLAM)~\cite{fu2021fast}. Nowadays, increasing attention has been attracted onto the multimodal feature extraction and matching in special scenarios, such as autonomous drive and remote sensing, because different modalities provide complementary information~\cite{zhang2021image}. Although several modal-invariant features,~\eg, OS-SIFT~\cite{xiang2018sift} and RIFT~\cite{li2018rift} emerge endlessly, the role SIFT~\cite{lowe2004distinctive} playing in visual image matching cannot be found for multimodal images. Therefore, it is imperative to study a more general and robust solution.

Modeling invariance is the key to feature extraction~\cite{jiang2021review}. Benefiting from the great potential of Deep Neural Network (DNN), the features learned on big data dispense with heuristic designs to acquire invariance and significantly outperform their traditional counterparts on both visual-only~\cite{mishchuk2017working,zhang2019learning,tian2019sosnet,tian2020hynet,ma2021sdgmnet,dusmanu2019d2,revaud2019r2d2,tyszkiewicz2020disk} and cross-modal~\cite{aguilera2017cross,liu2018h,cui2021map,cui2021cross} images. Deep learning methods can be mainly divided into two categories: the two-stage and the one-stage frameworks. The efforts~\cite{mishchuk2017working,tian2019sosnet,ma2021sdgmnet,aguilera2017cross,liu2018h} belonging to the former category are based on the manual detector and then encode the patches centered at detected interest points with DNN. Undoubtedly, those descriptors are limited by the detected scale, orientation and so on. To fill the gap between the detection and the description, the one-stage pipeline~\cite{detone2018superpoint,dusmanu2019d2,revaud2019r2d2,bhowmik2020reinforced,tyszkiewicz2020disk,liu2021dgd,cui2021cross,cui2021map,wang2021local,shen2022learning} that learns to output dense detection scores and descriptors is proposed and further improvements are achieved.

The joint framework seems alluring, however, its training would be unstable without a proper definition of detection. To address the problem, SuperPoint~\cite{detone2018superpoint} generates synthetic corner data to give the detection clear supervision. A SIFT-like non-maximum suppression is performed on the final feature map in D2-Net~\cite{dusmanu2019d2}. R2D2~\cite{revaud2019r2d2} proposes a slack constraint, in which detection scores are encouraged to peak locally and repeat among various images so that more dense points can be detected. Furthermore, to increase the reliability of the detected features, the detection is always coupled with the description in the optimization. For example, D2-Net tries to suppress the detection scores of the descriptors that are not distinct enough to match. Similarly, R2D2 learns an extra reliability mask to filter out those points. Additionally, the probabilistic model introduced in ReinforcedPoint~\cite{bhowmik2020reinforced} and DISK~\cite{tyszkiewicz2020disk} can be also seen as a coupling strategy that shares the same motivation with D2-Net and R2D2.

Compared with the synthetic supervision and the non-maximum suppression, the constraints of local peaking and repeatability are more feasible for detection, because of their flexibility in the training and practical significance in the test. Based on these properties, the detection scores should be also linked to the probability of correctly matching of corresponding descriptors,~\ie, the detection should be coupled with the description as mentioned above. However, the modal-invariant descriptors are always hard to learn and match. Naive suppression on the detected probability of those descriptors that are likely to be wrongly matched would fall into the local minimum where the detected probabilities are all zeros. Additionally, those hard descriptors are the key to gaining promotions, so simply ignoring them would not be a wise choice. Therefore, the coupling of detection and description should be more cautiously designed.

\begin{figure}
	\centering
	\includegraphics[width=0.95\linewidth]{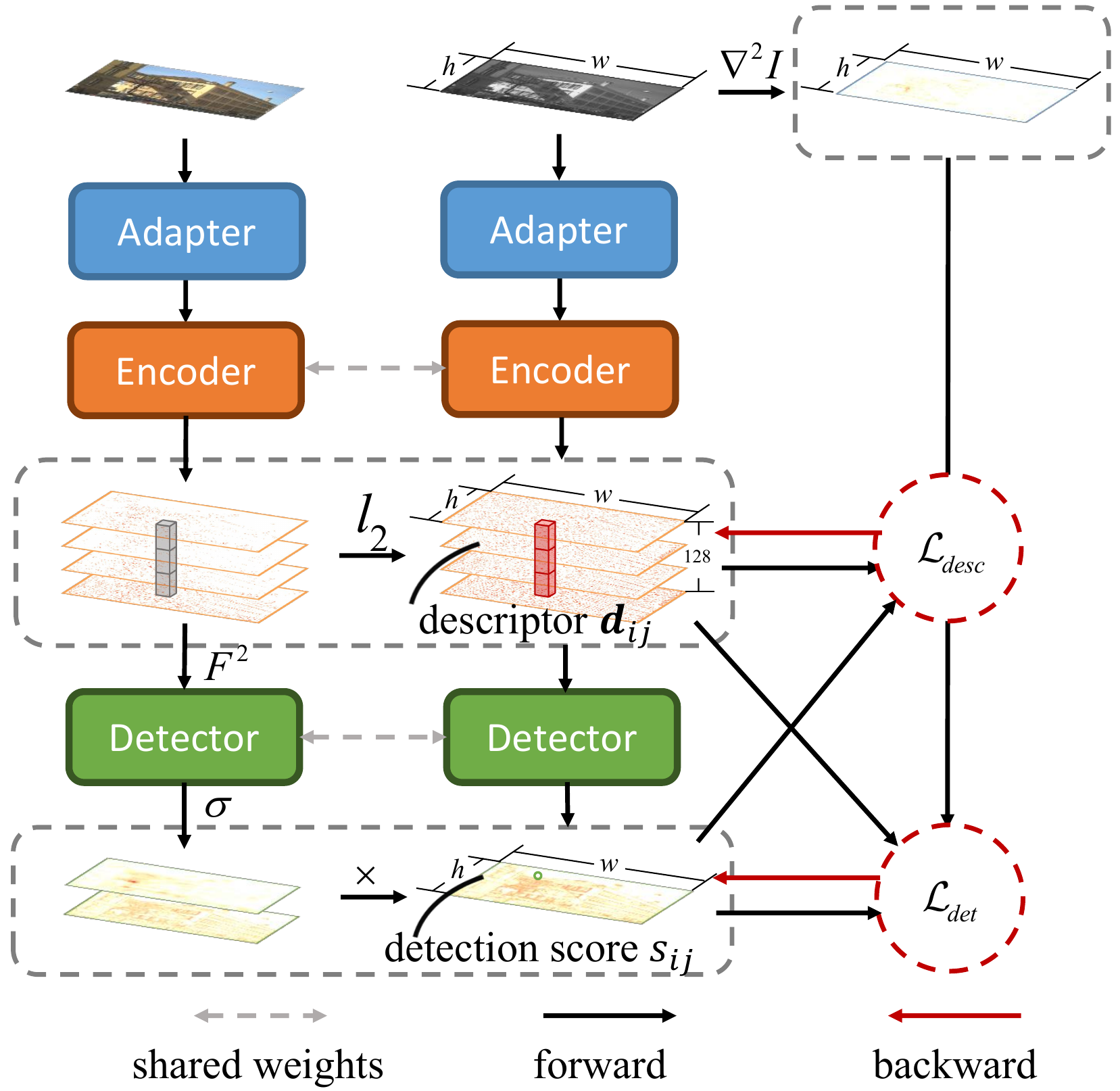}
	\caption{Overview of RedFeat, where $l_2$ denotes $l_2$ normalization and $\sigma$ denotes the activation function. In our method, an image successively passes through the modal-specific adapter, the weight-sharing encoder and the detector to generate a dense feature map and a detection score map. The basic loss function of descriptors,~\ie, features, is weighted with the corresponding detection scores. Meanwhile, detection scores are encouraged to peak according to the reliability of the corresponding descriptors and fade in smooth areas. Note that, no back-propagated flow traces back along the crossed forward flows that carry only the detached weights, which is the main idea of our recoupling strategy.  }
	
	\label{fig:fig1}
\end{figure}

In this paper, we firstly absorb the experience from related works and reformulate independent basic loss functions that are more effective and stable for multimodal feature learning, including a contrastive loss for description, a local peaking loss and a repeatability loss for detection. Different from the direct multiplication in previous efforts, we recouple the detection and the description in the mutual weighting strategy as briefly illustrated in Fig.~\ref{fig:fig1}. As for the detection, while an edge-based priori guides the detector to pay attention around edges, the detection scores of the reliable descriptors are further forced to peak by weighting the peaking loss with the matching risk of descriptors. Moreover, the repeatability loss is weighted by the similarity of corresponding descriptors. As for the description, the constrictive loss is weighted with the detection scores so that those descriptors with high detected probability are prioritized in the optimization. Note that, the weights in our recoupling strategy are `stopped gradients',~\ie, detached from back propagation, which makes the detection and the description would not be disturbed by the gradients of the weights. Finally, the features constrained by the recoupled detection and description loss functions, named ReDFeat, can be readily trained from scratch.

Moreover, to fulfill those harsh terms of the detection, Super Detector is proposed. The prior probability that a point is a keypoint and the conditional probability that a keypoint is extracted from the image are modeled by a fully connected network and a deep convolutional network with a large receptive field, respectively. Particularly, the deep convolution network is equipped with learnable local non-maximum suppression layers to stick out the keypoints. Finally, the posterior detected probability that a detected point is a keypoint is computed by the multiplication of outputs from two networks. To evaluate the features systematically, we collect several kinds of cross-modal image pairs, including cross visible (VIS), near-infrared (NIR), infrared (IR) and synthetic aperture radar (SAR), and build a benchmark in which the performances of features are strictly evaluated. Extensive experiments on this benchmark confirm the applicability of our ReDFeat.

Our contributions can be summarized as follows:
\begin{itemize}
	\item[1)] We recouple detection and description with the mutual weighting strategy, which can increase the training stability and performance of the learned cross-modal features.
	
	\item[2)] We propose Super Detector that possesses a large receptive field and learnable local non-maximum suppression blocks to improve the ability and the discreteness of the detection.
	
	\item[3)] We build a benchmark that contains three kinds of cross-modal image pairs for multimodal feature learning. Extensive experiments on the benchmark demonstrate the superiority of our method.

\end{itemize}

\section{Related Works}
\textbf{Handcrafted Methods.} Although visible deep learning features have sprung up in recent years, their handcrafted counterparts, such as SIFT~\cite{lowe2004distinctive}, ORB~\cite{mur2015orb} and SURF~\cite{bay2006surf} still maintain popular in common scenes due to their robustness and cheapness~\cite{jin2021image,jiang2021review}. Their cross-modal counterparts, such as MFP~\cite{aguilera2012multispectral}, SR-SIFT~\cite{sedaghat2015distinctive}, PCSD~\cite{fan2018sar}, OS-SIFT~\cite{xiang2018sift}, RIFT~\cite{li2018rift} and KAZE-SAR~\cite{pourfard2021kaze}, still receive a large number of attention from the community of multimodal image processing due to the scarcity of well registered data that can support the deep learning methods. Both the visible and the handcrafted multimodal features focus on corner or edge detection and description, which are believed to contain information that is invariant to geometric and radiant distortion. Their successes motivate many deep learning methods~\cite{detone2018superpoint,liu2021dgd,barroso2019key} and us to inject edge-based priori into the training.

\textbf{Two-stage Deep Methods.} In recent years, the deep learning `revolution' has swept across the whole field of computer vision including local feature extraction. However, due to the lack of strong supervision of detection, deep local descriptors~\cite{mishchuk2017working,tian2019sosnet,ma2021sdgmnet,aguilera2017cross,liu2018h} had been stuck in a two-stage pipeline, in which the keypoints are extracted by classical detectors,~\eg, Difference of Gaussian (DOG)~\cite{lowe2004distinctive}, and then patches centered in those points are encoded into descriptors by DNN. This pipeline restricts the room for modifications so that most methods devote to modifying the loss functions for descriptors~\cite{tian2019sosnet,zhang2019learning,tian2020hynet,ma2021sdgmnet}. Additionally, Key.Net~\cite{barroso2019key} takes the early effort to learn keypoint extraction with repeatability as a constraint, which is a minority of two-stage methods. Despite of the isolation between detection and description in this kind of methods, DNN still reveals strong potential for local feature extraction~\cite{jin2021image}. Moreover, the independent constraints of detection and description in the two-stage pipeline pave the way to the joint frameworks, which are also the bases of our formulation.

\textbf{One-stage Deep Methods.} An obvious limitation, that detection and description cannot reinforce each other, exists in the two-stage pipeline. To tackle this problem, SuperPoint~\cite{detone2018superpoint} firstly proposes a joint detection and description framework, in which the detection and the description are trained with the synthetic supervision and the contrastive learning, respectively.

To further enhance the interaction between these two steps, a framework for joint training with a semi-handcrafted local maximum mining as the detector,~\ie, D2-Net~\cite{dusmanu2019d2} is proposed. The detection and description of D2-Net are not only optimized in the meantime, but also tangled for the mutual guide. However, its non-maximum suppression detector is unexplainable. Based on D2-Net, SAFeat~\cite{shen2022learning} designs a multi-scale fusion network to extract scale-invariant features. CMMNet~\cite{cui2021cross} applies D2-Net to multimodal scenario. So, both SAFeat and CMMNet inherit the weakness of D2Net.

R2D2~\cite{revaud2019r2d2} proposes a fully learnable detector with feasible constraints and further introduces an extra learnable mask to filter out some unreliable detection. But it is unclear why the reliability should be additionally learned rather than fused into one detector. TransFeat~\cite{wang2021local} introduces transformer~\cite{vaswani2017attention} to capture global information for local feature learning. It has been aware of the flaw of D2-Net's detection and drawn on the local peaking loss from R2D2 to remedy the fault.

Furthermore, ReinforcedPoint~\cite{bhowmik2020reinforced} and DISK~\cite{tyszkiewicz2020disk}, model the matching probability in a differential form, in which the detection and description are concisely coupled, and employ Reinforcement Learning (RL) to construct and optimize the matching loss. Undoubtedly, the matching performance of DISK would significantly benefit from direct optimization on matching loss, but RL is hungry for computation and data, which might not be feasible in multimodal scenario.

To the best of our knowledge, besides CMMNet, MAP-Net~\cite{cui2021map} is the only joint detection and description method customized for multimodal images. However, it draws on the pipeline from DELG~\cite{cao2020unifying}, whose features are specific for the image retrieval task instead of the accurate image matching task that we focus on. Therefore, we tend to conduct a further study on more feasible joint detection and description methods for cross-modal image matching.

\section{Method}
\subsection{Background of Coupled Constraints}
\label{sec:sec3.1}
The joint detection and description framework of local feature learning aims to employ DNN to extract a dense descriptor map $\boldsymbol{D}(\boldsymbol{\Omega_a},\boldsymbol{\Omega_e}) \in \mathbb{R}^{H \times W\times C}$ and a detection score map $\boldsymbol{S}(\boldsymbol{\Omega_a},\boldsymbol{\Omega_e}, \boldsymbol{\Omega_d}) \in \mathbb{R}^{H \times W}$ for an input image $\boldsymbol{I} \in \mathbb{R}^{H \times W}$, where $\boldsymbol{\Omega_a}$, $\boldsymbol{\Omega_e}$, $\boldsymbol{\Omega_d}$ denote the parameters of adapter, encoder and detector, respectively. Let $\{i,i'\}$ represent a correspondence in a pair of overlapping images $\boldsymbol{I}$ and $\boldsymbol{I}'$, $\boldsymbol{d}_{i} \in \boldsymbol{D}$ represent the descriptor of $i$th point with detected probability $s_i\in \boldsymbol{S}$.

To constrain the learning of an individual descriptor $\boldsymbol{d}_{i}$, a matching risk function $\mathcal{R}_i(\boldsymbol{d}_i;\boldsymbol{D},\boldsymbol{D}')$ is constructed within descriptor maps $\boldsymbol{D}$ and $\boldsymbol{D}'$. Since the reliability of the descriptor can be estimated by $\mathcal{R}_i(\boldsymbol{d}_i;\boldsymbol{D},\boldsymbol{D}')$ to some extent, many related works couple the corresponding detection scores $s_i$ and $s_{i'}$ to guide the optimization of $\mathcal{R}_i(\boldsymbol{d}_i;\boldsymbol{D},\boldsymbol{D}')$ in a general loss:

\begin{equation}
	\mathcal{L}_g(\boldsymbol{D},\boldsymbol{D}',\boldsymbol{S},\boldsymbol{S}')=\mathbb{E}_i(
	s_is_{i'}\mathcal{R}(\boldsymbol{d}_i;\boldsymbol{D},\boldsymbol{D}')),
	\label{eqn:eqn1}
\end{equation}
where $\mathbb{E}(\cdotB)$ denotes the expectation calculation (averaging in batch). While the descriptors with larger detection score $s_i$ would play more important roles during the optimization, the detection scores of points with large $\mathcal{R}(\boldsymbol{d}_i;\boldsymbol{D},\boldsymbol{D}')$ would tend to be zeros. In this way, problems come up: firstly, the zero detection score map is one of a local minimum of this loss, which is not our desire. Secondly, the hard descriptors with large $\mathcal{R}(\boldsymbol{d}_i;\boldsymbol{D},\boldsymbol{D}')$ are the key for improvement, so they deserve more attention instead of being treated as distractors. The two problems are magnified in multimodal feature learning, in which the correspondences suffer from the extreme variance of imaging.

Although D2-Net~\cite{dusmanu2019d2} and CMMNet~\cite{cui2021cross} conduct normalization onto detection score $s$ so that $\boldsymbol{S}=0$ would not be the minimum of the loss, the normalization breaks the convexity of detection and the balance between the learning of detection and description would be hard to hold, which finally leads to failed optimization. Reliability loss of R2D2~\cite{revaud2019r2d2} also contains a term similar to Eqn.~\eqref{eqn:eqn1}, which would suffer from the first problem mentioned before. The failures of CMMNet and R2D2 prove our hypothesis as shown in Section~\ref{sec:sec5}. Moreover, the formulation of probabilistic matching loss introduced in ReinforcedPoint~\cite{bhowmik2020reinforced} and DISK~\cite{tyszkiewicz2020disk} is also similar to Eqn.~\eqref{eqn:eqn1}, so it is likely to get stuck in the two problems. Therefore, we devote ourselves to recoupling detection and description in a more elaborated way for better training.

\subsection{Basic Constraints}
The basic constraints of detection and description should be determined before coupling them. To satisfy the nearest neighbor matching principle, the distance between an anchor and its nearest non-matching neighbor should be maximized, while the distance to its correspondence should be minimized. Therefore, we sample $N$ pairs of corresponding descriptors $\{\boldsymbol{d}_i,\boldsymbol{d}_{i'}\}$ and their detection scores $\{s_i,s_{i'}\}$. The set of samples is denoted by $\{\boldsymbol{D}_N,\boldsymbol{D}'_N\}$. For a pair corresponding cross-modal descriptors $\{\boldsymbol{d}_i,\boldsymbol{d}_{i'}\}$, we mine two intra-modal nearest non-matching neighbor $\{\boldsymbol{d}_j,\boldsymbol{d}_{k}\}$ and two inter-modal non-matching neighbor $\{\boldsymbol{d}_m,\boldsymbol{d}_n\}$ in $\{\boldsymbol{d}_i,\boldsymbol{d}_{i'}\}$ as:
\begin{align}
	\label{eqn:eqn2}
	\boldsymbol{d}_j&=\text{argmin}_{\boldsymbol{d}_j\in\boldsymbol{D}_N,j\ne i} \theta(\boldsymbol{d}_i,\boldsymbol{d}_j),\\
	\label{eqn:eqn3}
	\boldsymbol{d}_k&=\text{argmin}_{\boldsymbol{d}_k\in\boldsymbol{D}'_N,k\ne i'} \theta(\boldsymbol{d}_{i'},\boldsymbol{d}_k),\\
	\label{eqn:eqn4}
	\boldsymbol{d}_n&=\text{argmin}_{\boldsymbol{d}_n\in\boldsymbol{D}'_N,n\ne i'} \theta(\boldsymbol{d}_{i},\boldsymbol{d}_n),\\
	\label{eqn:eqn5}
	\boldsymbol{d}_m&=\text{argmin}_{\boldsymbol{d}_m\in\boldsymbol{D}_N,m\ne i} \theta(\boldsymbol{d}_{i'},\boldsymbol{d}_m),
\end{align}
where $\theta(\boldsymbol{d}_i,\boldsymbol{d}_j)=\text{acos}(\boldsymbol{d}_i^T\boldsymbol{d}_j)$ is the angular distance. Although the nearest neighbor matching request distinction between an anchor and its inter-modal nearest neighbor, we believe maximizing $\theta(\boldsymbol{d}_i,\boldsymbol{d}_n)$ and $\theta(\boldsymbol{d}_i,\boldsymbol{d}_m)$ in the meantime is hazardous for acquiring the modal invariance. Thus we tend to maximize $\text{max}\{\theta(\boldsymbol{d}_i,\boldsymbol{d}_n),\theta(\boldsymbol{d}_i,\boldsymbol{d}_m))\}$, $\theta(\boldsymbol{d}_i,\boldsymbol{d}_k)$ and $\theta(\boldsymbol{d}_i,\boldsymbol{d}_j)$, while minimizing $\theta(\boldsymbol{d}_i,\boldsymbol{d}_{i'})$ in contrastive learning behavior. As a result, our basic loss function of description $\mathcal{L}_\text{desc-B}$ is:
\begin{align}
	\begin{split}
	\mathcal{R}(\boldsymbol{d}_i;\boldsymbol{D}_N,\boldsymbol{D}'_N)= &[(\pi-\theta(\boldsymbol{d}_i,\boldsymbol{d}_k))^2+(\pi-\theta(\boldsymbol{d}_i,\boldsymbol{d}_j))^2\\
	&+(\pi-\text{max}\{\theta(\boldsymbol{d}_i,\boldsymbol{d}_n),\theta(\boldsymbol{d}_i,\boldsymbol{d}_m) \})^2\\
	&+3\theta(\boldsymbol{d}_i,\boldsymbol{d}_{i'})^2 ]^2,
	\end{split}
	\\
	\mathcal{L}_{\text{desc-B}}(\boldsymbol{D}_N,\boldsymbol{D}'_N)=& \Expect_i(\mathcal{R}(\boldsymbol{d}_i;\boldsymbol{D}_N,\boldsymbol{D}'_N)).
\end{align}
The angular distance is employed for distance measure because it could balance the optimization of matching and non-matching pairs~\cite{ma2021sdgmnet}. Moreover, the quadratic matching risk makes the hard samples obtain larger gradients to be optimized. In this way, $\mathcal{L}_{\text{desc-b}}$ is expected to increase the cross-modal robustness of descriptors.

As mentioned above, repeatability and local peaking should be the primary properties of detection. To guarantee the repeatability, the detection score $\boldsymbol{S}$ of the first image should be similar to the warped $\boldsymbol{S}'_w$ of the other image. Moreover, the detection score should be salient so that a unique point can be extracted in a local area. Thus, we follow R2D2~\cite{revaud2019r2d2} to primarily constrain the detection with basic repeatability loss $\mathcal{L}_{\text{rep-B}}$ and peaking loss $\mathcal{L}_{\text{peak-B}}$ as:
\begin{align}
	\mathcal{L}_{\text{rep-B}}(\boldsymbol{S},\boldsymbol{S}'_w)&=\mathbb{E}_p(1-\boldsymbol{S}[p]^T\boldsymbol{S}'_w[p]),
	\\
	\label{eqn:eqn9}
	\mathcal{L}_{\text{peak-B}}(\boldsymbol{S})&=\mathbb{E}_i(\mathrm{AP}(\boldsymbol{S})[i]^2+(1-\mathrm{MP}(\boldsymbol{S}))[i])^2,
\end{align}
where $p$ is a flattened patch of coordinate, which is extracted on full coordinate $\{1,\dots,H\}\times\{1,\dots,W\}$ by shifted windows with kernel size of $17\times 17$ and stride of $8$; $\boldsymbol{S}[p]\in \mathbb{R}^{256}$ denotes the flattened and normalized vector of detection score, which is indexed by $p$; AP and MP denote the average pooling and the max pooling with kernel size of $17\times17$ and stride of $1$, respectively. Note that, the kernel size and the strides are all adopted from R2D2 empirically.

\subsection{Recoupled Constraints}
\begin{figure}[]
	\centering
	\includegraphics[width=0.9\linewidth]{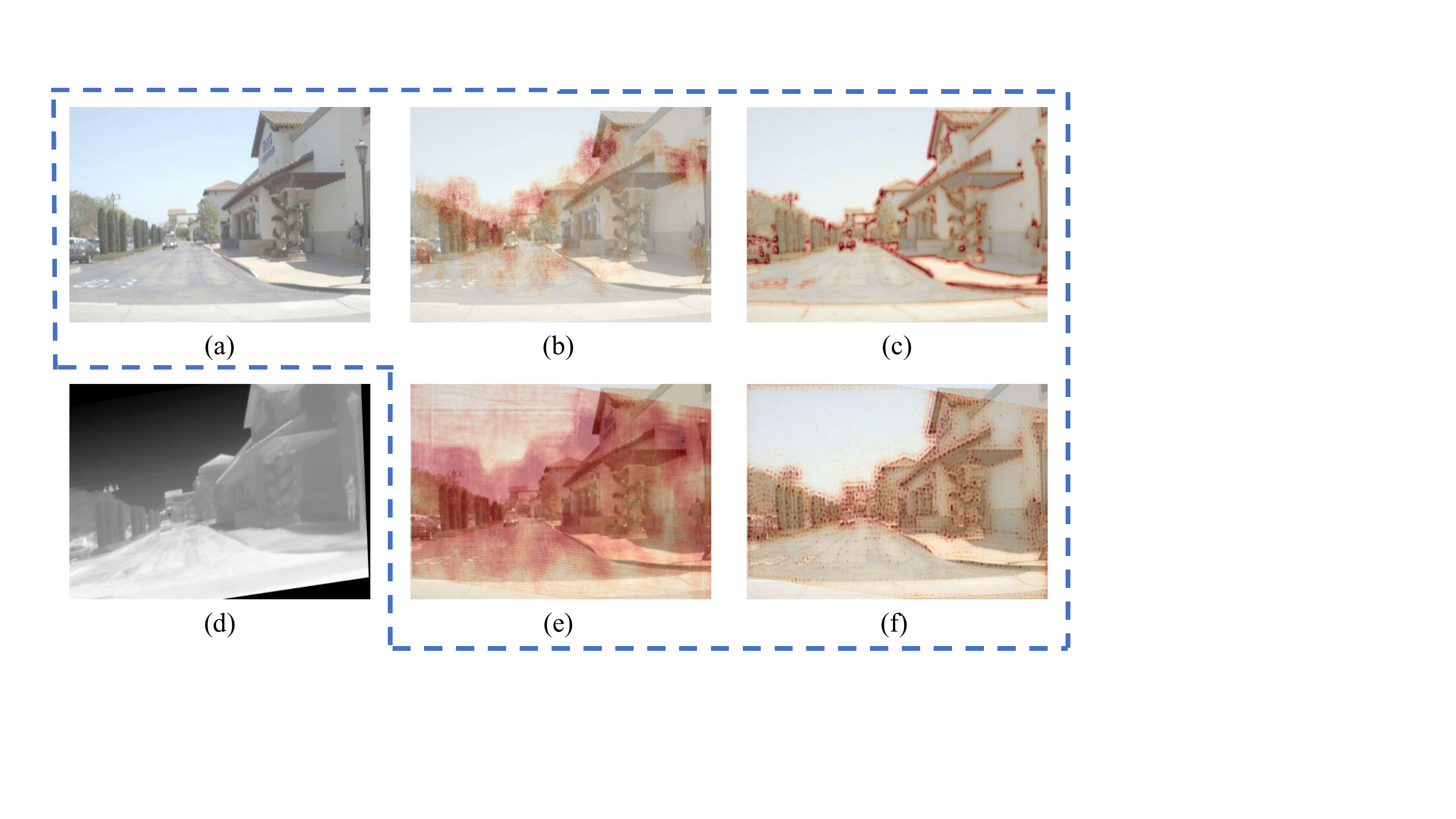}
	\caption{Visualization of weights of the input image pair (a) (d) in our recoupled constraints, where the darker red indicates the larger relative value. (a) The input visible image. (b) Visualization of $a(\mathcal{R}(\boldsymbol{d}_i;\boldsymbol{D_N},\boldsymbol{D_N}'))$, in which batches of descriptors are randomly sampled. (c) The edge of (a) detected by Laplacian operator. (d) The warped infrared image. (e) Similarity of corresponding descriptors used to weight the repeatability learning. (f) visualization of $c(s_i,s_i')$ employed for the guided descriptor learning.  }
	\label{fig:fig2}
\end{figure}

Successes of the related works \cite{dusmanu2019d2,revaud2019r2d2,tyszkiewicz2020disk} suggest that coupling detection and description can improve the feature learning, however, inappropriate coupling strategies bring up problems as mentioned in Section~\ref{sec:sec3.1}. To tackle problems, we recouple them with a mutual weighting strategy, in which the gradients of weights are `stopped' as illustrated in Fig.~\ref{fig:fig1}. Specifically, we again sample $N$ pair of corresponding descriptors $\{\boldsymbol{d}_i,\boldsymbol{d}_{i'}\}$ and their detection scores $\{s_i,s_{i'}\}$. For detection, a weight ${a}(\mathcal{R}(\boldsymbol{d}_i;\boldsymbol{D},\boldsymbol{D}')$ that is negatively correlated to matching risk $\mathcal{R}(\boldsymbol{d}_i;\boldsymbol{D}_N,\boldsymbol{D}'_N)$ would encourage the more reliable descriptor to be detected in a higher probability. Furthermore, learning from the handcrafted cross-modal features which focus on modal-invariant texture extraction, we introduce an edge-based prior $\boldsymbol{M}(\boldsymbol{I})$ to prevent the interest points from laying on smooth areas. So the recoupled peaking loss $\mathcal{L}_{\text{peak-R}}$ can be formulated as:
\begin{align}
	{a}(\mathcal{R}(\boldsymbol{d}_i;\boldsymbol{D},\boldsymbol{D}')\triangleq\ &\bigg[1-\frac{R(\boldsymbol{d}_i;\boldsymbol{D},\boldsymbol{D}')}{\mathbb{E}_i(R(\boldsymbol{d}_i;\boldsymbol{D},\boldsymbol{D}'))}\bigg]_+,\\
	\boldsymbol{M}(\boldsymbol{I})=\ &\bigg[1-\frac{\|\nabla^2\boldsymbol{I}\|}{\mathbb{E}_i(\|\nabla^2\boldsymbol{I}\|[i])}\bigg]_+,\\
	\begin{split}
	\label{eqn:eqn12}
	\mathcal{L}_{\text{peak-R}}(\boldsymbol{S},\boldsymbol{D},\boldsymbol{D}',\boldsymbol{I})=\ &\mathcal{L}_{\text{peak-B}}(\boldsymbol{S})+\mathbb{E}_i((\boldsymbol{M}(\boldsymbol{I})\boldsymbol{S})[i]^2)+\\
	&\mathbb{E}_i(a(\mathcal{R}(\boldsymbol{d}_i;\boldsymbol{D}_N,\boldsymbol{D}_N'))(1-s_i)^2),
	\end{split}
\end{align}
where $[\cdot]_+$ denotes the rectified linear unit (ReLU); $\triangleq$ denotes the `stop gradient equality'; the edge of image $\boldsymbol{I}$ is computed as $\|\nabla^2\boldsymbol{I}\|$ with Laplacian operator. The weights ${a}(\mathcal{R}(\boldsymbol{d}_i;\boldsymbol{D},\boldsymbol{D}')$ and $\boldsymbol{M}(\boldsymbol{I})$ are visualized in Fig.~\ref{fig:fig2} (b) and (c), respectively.

There are several key differences between our peaking constraints and previous works. Firstly, the recoupling weight $a(\mathcal{R}(\boldsymbol{d}_i;\boldsymbol{D},\boldsymbol{D}'))$ is detached from back propagation, which would not directly affect the learning of description~\cite{dusmanu2019d2,cui2021cross,revaud2019r2d2,tyszkiewicz2020disk}. Secondly, $a(\mathcal{R}(\boldsymbol{d}_i;\boldsymbol{D},\boldsymbol{D}'))$ only constrains the peaking of the detection, which would not suppress any detected probability of hard descriptors with large $\mathcal{R}(\boldsymbol{d}_i;\boldsymbol{D},\boldsymbol{D}')$ and solve the problems mentioned above~\cite{dusmanu2019d2,cui2021cross,revaud2019r2d2,tyszkiewicz2020disk}. Thirdly, edge-based priori $\boldsymbol{M}(\boldsymbol{I})$ is introduced to balance the peaking constraint, instead of forcing the model to detect corners or edges~\cite{detone2018superpoint,liu2021dgd,barroso2019key}. Moreover, the weights are normalized by the expectation dynamically, so the weights would not be zeros and keep functioning. In this way, while the detection turns explainable and reliable, it can be more stably trained without risks of falling into sub-optimum,~\ie, trivial solution.

Multimodal sensors may display the same object in totally different forms, which means requesting repeatability in such areas is exactly irrational. Thus, the repeatability also needs guides from the description. For two corresponding patches of detection score, we compute the average cosine similarity of their descriptors to estimate the local similarity. Then, the local similarity is used as a weight to modulate the recoupled repeatability loss as:
\begin{align}
	b(p;\boldsymbol{D},\boldsymbol{D}_w')\triangleq\ &\mathbb{E}_{i}(\boldsymbol{D}[p[i]]^T\boldsymbol{D}'_w[p[i]]), \\
	\label{eqn:eqn14}
	\begin{split}
			\mathcal{L}_{\text{rep-R}}(\boldsymbol{S},\boldsymbol{S}'_w)=\ &\mathbb{E}_p(	b(p;\boldsymbol{D},\boldsymbol{D}_w')(1-\\
			&\boldsymbol{S}[p]^T\boldsymbol{S}'_w[p]))),
	\end{split}
\end{align}
where $\boldsymbol{D}_w'$ denotes the warped dense descriptors. Note that the detached $b(p;\boldsymbol{D},\boldsymbol{D}_w')$ would also not affect the optimization of description. And $\boldsymbol{D}[p[i]]^T\boldsymbol{D}'_w[p[i]]$ in the weight $b(p;\boldsymbol{D},\boldsymbol{D}_w')$ is visualized in Fig.~\ref{fig:fig2} (e).

As discussed before, since the flattening and peaking of detection are safely defined and hold a balance in the recoulped peaking loss, the detection would slip into trivial solution,~\eg, zeros. Therefore, it is worth recoupling the detection to the description. In other words, the matching risk can be weighted with the detection score so that the descriptors with high detected probability would attract more attention in the optimization. The recoupled description loss with detached weights can be formulated as:
\begin{align}
	c(s_i,s_i')\triangleq\ &s_is_i', \\
	\mathcal{L}_{\text{desc-R}}(\boldsymbol{D}_N,\boldsymbol{D}'_N)=\ &\Expect_i(c(s_i,s_i))\mathcal{R}(\boldsymbol{d}_i;\boldsymbol{D}_N,\boldsymbol{D}'_N)),
\end{align}
where $c(s_i,s_i')$ is shown in Fig.~\ref{fig:fig1} (f).

Finally, the total loss function of our RedFeat can be formulated in the sum of Eqns.~\eqref{eqn:eqn9},~\eqref{eqn:eqn12} and~\eqref{eqn:eqn14} with only one hyperparameter $\lambda$:
\begin{equation}
	\begin{aligned}
		\mathcal{L}=\ &\mathcal{L}_{\text{desc-R}}(\boldsymbol{D}_N,\boldsymbol{D}'_N)+\mathcal{L}_{\text{peak-R}}(\boldsymbol{S},\boldsymbol{D},\boldsymbol{D}',\boldsymbol{I})+\\
		&\mathcal{L}_{\text{peak-R}}(\boldsymbol{S}',\boldsymbol{D}',\boldsymbol{D},\boldsymbol{I}')+\lambda \mathcal{L}_{\text{rep-R}}(\boldsymbol{S},\boldsymbol{S}'_w).
	\end{aligned}
\end{equation}
While the weights ${a}(\mathcal{R}(\boldsymbol{d}_i;\boldsymbol{D},\boldsymbol{D}')$ and $b(p;\boldsymbol{D},\boldsymbol{D}_w')$ are generated by description and recoulped to the detection, the weight $c(s_i,s_i')$ takes a converse effect in the loss. This loss based on the mutual weighting strategy would stabilize and boost the feature learning.

\subsection{Network Architecture}
\textbf{Architecture}. Most joint detection and description methods share similar architectures which include an encoder and a detector. R2D2 proposes a lightweight encoder that contains only $9$ convolution layers and a naive linear detector to output 128-dimensional dense descriptors and a score map, which is cheap in time and memory. Therefore, we adopt this architecture as our raw architecture. The $6$ shallow layers are divided as the adapter that is unshared for eliminating the variance of modals. The raw encoder in our architecture consists of the last $3$ convolutional layers and the raw detector keeps the same with R2D2. Note that, the encoder and detector are weight sharing.

\textbf{Super Detector}. Our recoupling constraints mainly embrace the detection. limited by the small receptive field, the raw linear detector cannot capture the neighborhood and global information to fulfill peaking loss. Therefore, we propose a super detector, which has two branches like R2D2. One branch is the raw detector $\theta_{d0}$ that models the prior probability $p({kp}_i|\boldsymbol{d}_i)$ that the point $i$ is a keypoint as $p({kp}_i|\boldsymbol{d}_i)=\text{Sigmoid}(\theta_{d0}(\boldsymbol{d}_i))$; The structure of the other branch needs to model the conditional probability $p(dp_i|\boldsymbol{D},kp_i)$ that a keypoint can be detected globally.

\begin{figure}[]
	\centering
	\includegraphics[width=0.6\linewidth]{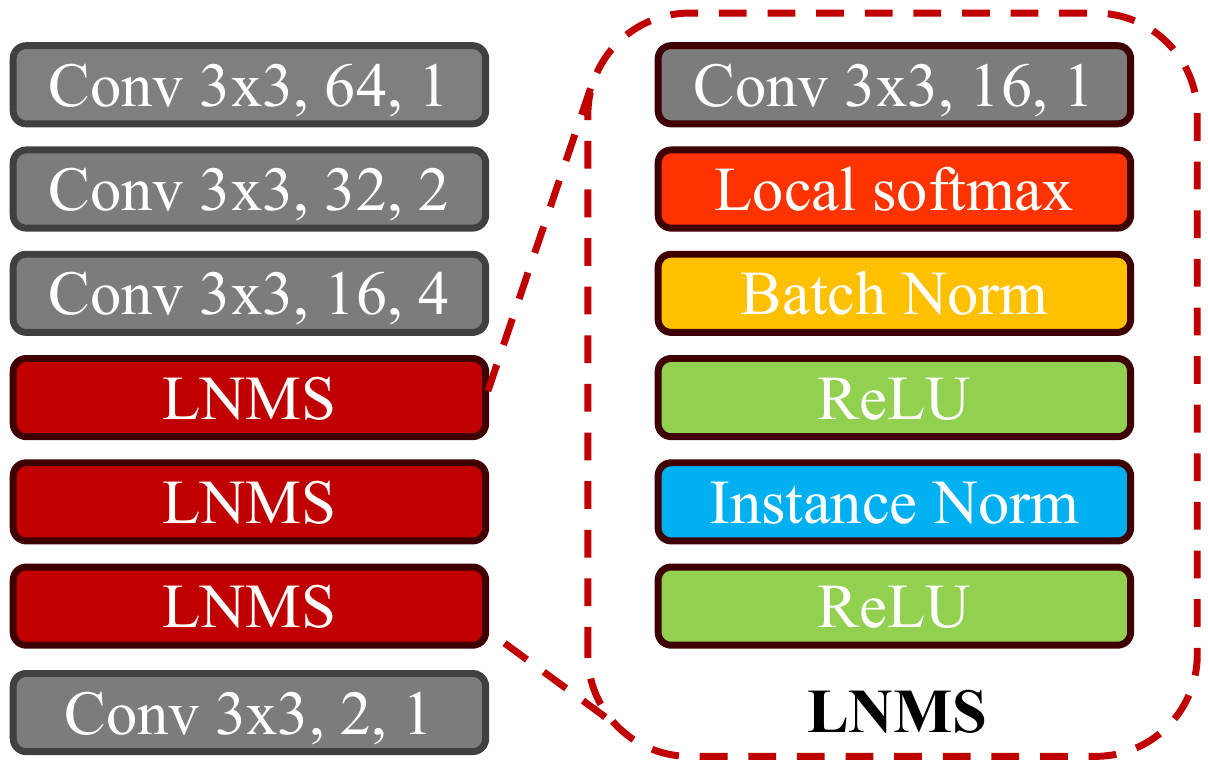}
	\caption{The branch for conditional probability estimation. Conv $3\times3$, $64$, $1$ denote convolutional layer with kernel size of $3\times3$, output channel of $64$ and dilation of $1$. Local softmax in learnable non-maximum suppression block is formulated as Eqn.~\eqref{eqn:eqn18}.}
	\label{fig:fig3}
\end{figure}

Since $p({dp}_i|\boldsymbol{D},{kp}_i)$ is related to global information, the branch should possess a larger receptive field by stacking more convolutional layers. Moreover, the score of the detected point should be the local maximum, so we propose learnable non-maximum suppression layer (LNMS) as shown in Fig.~\ref{fig:fig3}. In the LMNS, features are firstly transformed by a learnable convolutional layer. Then, local maximums in the transformed feature map are detected by the local softmax operation,~\ie, Eqn.~\eqref{eqn:eqn18}. At last, statistical maximums in batch and channel are further mined by BN and IN with ReLU. Briefly, for an input feature map $\boldsymbol{x}$, the forward propagation in LNMS can be described as:
\begin{align}
	\nonumber
	\boldsymbol{x} &= \text{Conv}(\boldsymbol{x}),\\
	\label{eqn:eqn18}
	\boldsymbol{x} &= \text{Exp}(\boldsymbol{x})/\text{AP3}(\text{Exp}(\boldsymbol{x})),\\
	\nonumber
	\boldsymbol{x} &= \text{ReLU}(\text{BN}(\boldsymbol{x})),\\
	\nonumber
	\boldsymbol{x} &= \text{ReLU}(\text{IN}(\boldsymbol{x})),
\end{align}
where AP3 denotes average pooling with a kernel size of $3$. BN and IN represent batch normalization and instance normalization, respectively. Finally, the branch is constructed by cascading convolutional layers and several LNMS as shown in Fig.~\ref{fig:fig3}, and it outputs a two-channel feature. After channel softmax activation, the first channel in the final feature map is maintained as $p({dp}_i|\boldsymbol{D},{kp}_i)$. The posterior probability $p({kp}_i|\boldsymbol{D},{dp}_i)$ that the detected point is an interest point,~\ie, $s_i$, can be approximately computed as $p({kp}_i|\boldsymbol{d}_i)p({dp}_i|\boldsymbol{D},{kp}_i)$.

\section{Benchmark}
The lack of benchmark is one of the major reasons for the slow development of multimodal feature learning. Therefore, building a benchmark might be even more imperative than a robust algorithm. In this paper, we collect three kinds of cross-modal images, including RGB-NIR, VIS-IR and VIS-SAR to build a benchmark for cross-modal feature learning. The features can be evaluated in feature matching and image registration pipelines. Basic information of the collected data is shown Table~\ref{tab:data}.

\subsection{Dataset}
\textbf{VIS-NIR}. Visible and near-infrared (NIR) image pairs in the average size of $983\times686$ are collected from the RGB-NIR scene dataset~\cite{brown2011multi}. The dataset covers various scenes, including country, field, forest, indoor, mountain, old building, street, urban, and water. And most image pairs are photographed in special conditions and can be well registered. We randomly split the images from $9$ scenes into the training set and the test set with a ratio of $3:1$, which results in a training set of $345$ pairs of images. The ground truths of the test set are manually validated and filtered for more reliable evaluation. Finally, there are $128$ pairs of images left in the test set.

\textbf{VIS-IR}. We collect roughly registered 265 pairs of visible and long-wave infrared (IR) images in the average size of $533\times321$. $44$ static image pairs from RGB-LWIR~\cite{aguilera2015lghd} are mainly shot on buildings during the day. The other $221$ pairs of video frames come from RoadScene~\cite{xu2020aaai}, in which more complex objects,~\eg, cars and people, are captured both day and night. We randomly select $47$ images as the test set and leave the rest as the training set. Since the overlapping image pairs cannot be registered with the homography matrix due to the greatly varying depth of objects, we manually mark about $16$ landmarks per image pair for reprojection error estimation.

\textbf{VIS-SAR}. Optical-SAR~\cite{xiang2020automatic} provides aligned gray level and synthetic aperture radar image pairs in the uniform size of $512\times512$, which are remotely sensed by the satellite and cover field and town scenes. There are 2011 and 424 image pairs in the training set and test set, respectively. The dataset and its split are gathered into our benchmark without changes. Note that, it is hard to validate the ground truth or label landmarks for this subset due to the fuzziness of SAR images.

\begin{table}[]
	\setlength{\tabcolsep}{1.4mm}
	\renewcommand\arraystretch{1.5}
	\centering
	\caption{Basic information of subsets in our benchmark. The number, the number of channel, the size and the character of the collected images are reported.}
	\begin{tabular}{l|cc|cc|c|l}
		\hline
	\multirow{2}{*}{} & \multicolumn{2}{c|}{Number}       & \multicolumn{2}{c|}{Channel} & \multirow{2}{*}{Size} & \multicolumn{1}{c}{\multirow{2}{*}{Character}} \\ \cline{2-5}
	& \multicolumn{1}{c|}{Train} & Test & \multicolumn{1}{c|}{VIS} & * &                       & \multicolumn{1}{c}{}                           \\ \hline
	VIS-NIR           & \multicolumn{1}{c|}{345}      &128     & \multicolumn{1}{c|}{3}   & 1 & $983\times686$               & Multiple scenes                                \\ \hline
	VIS-IR            & \multicolumn{1}{c|}{221}      &47      & \multicolumn{1}{c|}{3}   & 1&
	$533\times321$                       & Road video at night                    \\ \hline
	VIS-SAR           & \multicolumn{1}{c|}{2011}      &424      & \multicolumn{1}{c|}{1}   & 1 & $512\times512$               & Satellite remote sensing                     \\ \hline
	\end{tabular}
	\label{tab:data}
\end{table}

\subsection{Evaluation Protocol}
\textbf{Random Transform}. Cross-modal features should carry both geometric and modal invariance. Thus, we generate homography transforms $\boldsymbol{H}$ by cascading random perspective with distortion scale $[0,0.2]$, random rotation $[-10^\circ,10^\circ]$ and random scaling $[0.8,1.0]$. And then, the transforms $\boldsymbol{H}$ are conducted on the aligned raw test set to generate the warped test set.

\textbf{Feature Matching}. To generate sufficient matches, the detected keypoints should be repeatable and the extracted descriptors should be robust. To evaluate the repeatability of the keypoints, we compute the number of correspondences (Corres) of image pairs and the repeatable rate (RR),~\ie, the ratio of the number of correspondences over the detected keypoints. Furthermore, we match the descriptors with the bidirectional nearest neighbor matching and calculate the number of correct matches. Following the definition in~\cite{detone2018superpoint,revaud2019r2d2}, we report matching score (MS), that the ratio of the number of correct matches over the number of the detected keypoints in the overlap, to evaluate the robustness of descriptors. Note that, the metrics are validated at different thresholds of pixels. And RR and MS are computed symmetrically across the pair of images and averaged.

\textbf{Image Registration}. The image registration is the destination of local feature learning. The matched features are used to estimate homography transform $\boldsymbol{H}'$ with RanSAC from OpenCV libraries, where the reprojection threshold is set to $10$ pixels and the iterations to $100$K. Since the ground-truth transform $\boldsymbol{H}$ is provided, we compute the reprojection error $RE_H$ as:
\begin{equation}
	\text{RE}_ {H}(\boldsymbol{H},\boldsymbol{H}') = \| \boldsymbol{H}[:]- \boldsymbol{H}'[:] \|,
\end{equation}
where $\boldsymbol{H}[:]$ denotes the flatten vector of $\boldsymbol{H}$. However, this metric would be not indicative for VIS-IR subset, because the raw test image pairs are not well aligned. Therefore, we introduce another method to estimate reprojection error with the landmarks as:
\begin{equation}
	\text{RE}_{M}(\boldsymbol{H}',\boldsymbol{M},\boldsymbol{M}') = \mathbb{E}_i\|\tau(\boldsymbol{m}_i,\boldsymbol{H}')-\boldsymbol{m}_{i'}\|,
\end{equation}
where $\boldsymbol{M}$ and $\boldsymbol{M}'$ denote the set of landmarks $\boldsymbol{m}$ on two images; $\tau(\boldsymbol{m}_i,\boldsymbol{H}')$ represents the reprojected point of $\boldsymbol{m}_i$. The registration is successful, if RE is smaller than a threshold. The successfully registered images (SR) are counted and the successful registration rate (SRR) is calculated on each subset.

\section{Experiments}
\subsection{Implementation}
We implement our ReDFeat in PyTorch~\cite{paszke2019pytorch} and Kornia~\cite{riba2020kornia}. The training data in a size of $192\times192$ is achieved by cropping, normalization and random perspective transform mentioned above. The network is trained in about 10000 iterations with a batch size of 2 on an NVIDIA RTX3090 GPU. Adam optimizer~\cite{kingma2014adam} with weight decay of $0.0005$ is employed to optimize the loss. Its learning rate is initialized at $0.001$ and decays to $0$ at the last epoch. The last checkpoint of training would be used for evaluation.

Our ReDFeat is compared to several counterparts in our benchmark, including SIFT~\cite{lowe2004distinctive}, RIFT~\cite{li2018rift}, DOG+HN~\cite{mishchuk2017working}, R2D2~\cite{revaud2019r2d2} and CMMNet~\cite{cui2021cross}. SIFT and RIFT are extracted with the open-source codes and default settings. HardNet and R2D2, which are deep learning features for visible images, are modified to multimodal scenario by specializing parameters of the first $6$ convolutional layers for individual modal images. CMMNet, which is not open-source, is implemented on the codebase of D2Net~\cite{dusmanu2019d2}.

\begin{table}[t]
	\centering
	\setlength{\tabcolsep}{2.0mm}
	\renewcommand\arraystretch{1.5}
	\caption{Training Stability of 1024 keypoints of joint detection and description methods on VIS-IR.}
	\begin{tabular}{l|c|c|c}
		\hline
		\# Matches (MS) & R2D2    & CMMNet  & ReDFeat  \\ \hline
		Pre-trained   & 36 (3\%)   & 42 (4\%)   & 171 (16\%) \\ \hline
		Scratch      & 0 (0\%)  & 1 (0\%) & 160 (15\%) \\ \hline
	\end{tabular}
	
	\label{tab:tab1}
\end{table}

\begin{figure*}[t]
	\centering
	\includegraphics[width = 3.3cm]{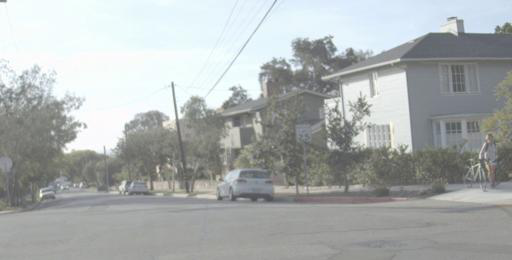}
	\includegraphics[width = 3.3cm]{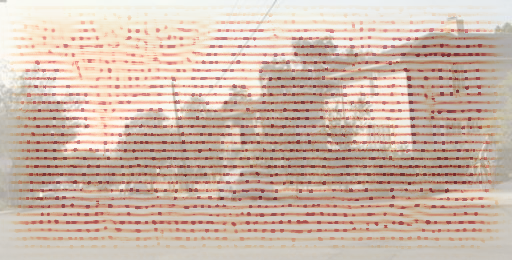}
	\includegraphics[width = 3.3cm]{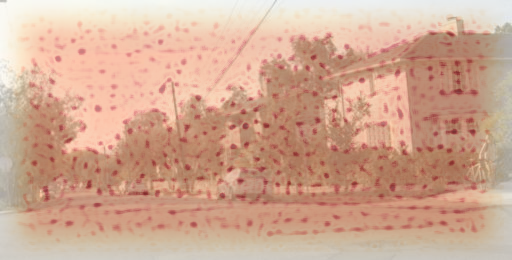}
	\includegraphics[width = 3.3cm]{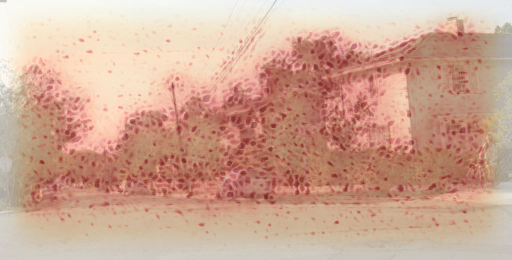}
	\includegraphics[width = 3.3cm]{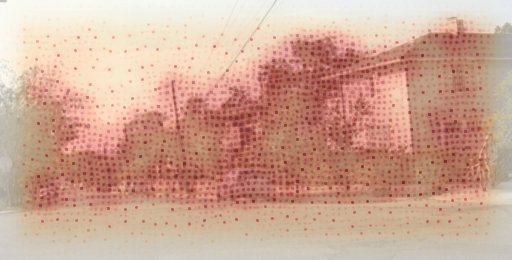}\\
	\vspace{0.05in}
	\includegraphics[width = 3.3cm]{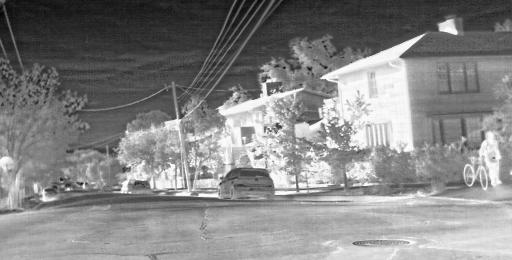}
	\includegraphics[width = 3.3cm]{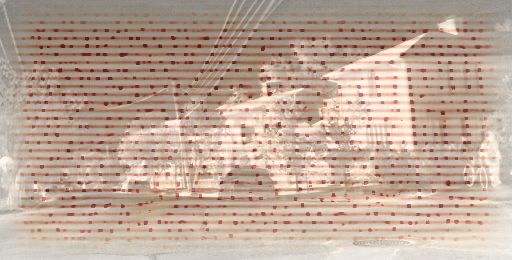}
	\includegraphics[width = 3.3cm]{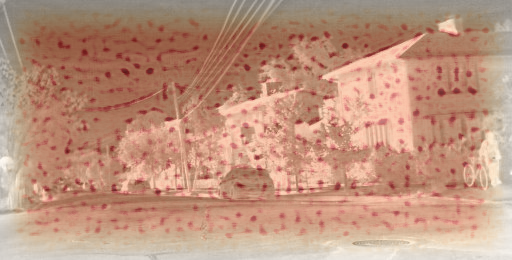}
	\includegraphics[width = 3.3cm]{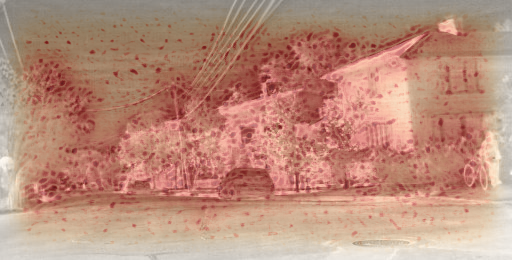}
	\includegraphics[width = 3.3cm]{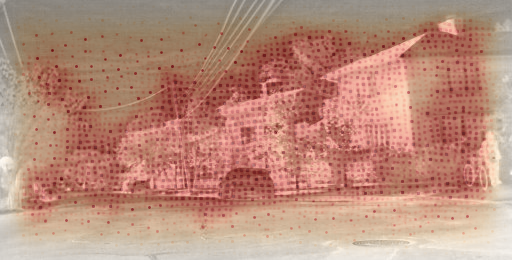}\\
	\raggedright
	{\footnotesize \hspace{5.3cm}  R2D2 \hspace{2.6cm} +{\textcircled{\scriptsize 1}} \hspace{2.6cm} +{\textcircled{\scriptsize 1}}+{\textcircled{\scriptsize 2}} \hspace{2cm} +{\textcircled{\scriptsize 1}}+{\textcircled{\scriptsize 2}}+{\textcircled{\scriptsize 3}}}
	{\small}
	\caption{Visualization of detection score in ablation study. R2D2 is chosen as the baseline and the others are our methods consisting of different proposed components. Darker red denotes the higher detected probability. The detection of R2D2 is computed by multiplying the repeatability and reliability.  }	
	\label{fig:abla}
\end{figure*}

\subsection{Ablation Study}
\textbf{Training Stability}. Since the training stability is the key problem that our recoupling strategy aims to tackle, we try to train the CMMNet, R2D2 and our method from scratch or pre-trained models to confirm our motivation. CMMNet adopts the VGG-16 pre-trained on ImageNet as the initialization. For comparison, we use the official pre-trained model for visible images to initialize R2D2 for cross-modal images. For ReDFeat, we just employ self-supervised learning~\cite{detone2018superpoint}, which is fed with augmented visible images, to obtain a pre-trained model. The mean number of correct matches and MS of $1024$ keypoints on VIS-IR subset are shown in Table~\ref{tab:tab1}. As we can see, CMMNet and R2D2 fail to learn discriminative features without pre-trained models, because the joint optimization of their naive coupled constraints is ill-posed. By contrast, our ReDFeat can be readily trained from scratch while also achieving a tiny improvement from the self-supervised pre-trained, which demonstrates the solidity of our formulation. Therefore, while we keep initializing the training of CMMNet and R2D2 with pre-trained models in subsequent experiments, our ReDFeat would be always trained from scratch.

\textbf{Impact of $\lambda$}. The weight of $\mathcal{L}_{\text{rep-R}}$, $\lambda$, is the only hyperparameter of ReDFeat, which plays a crucial role in balancing detection and description in our recoupling strategy. To investigate the impact of $\lambda$, we train ReDFeat with different $\lambda$ values and report relevant metrics of $1024$ keypoints on VIS-IR in Table~\ref{tab:tab2}. Totally, $\lambda$ lager than $8$ brings out desirable registration performance that the community focuses on. It demonstrates that the repeatability constrained by $\mathcal{L}_{\text{rep-R}}$ and weighted by $\lambda$ impose a strong impact on the registration performance. However, the repeatability not only forces the detection to be similar but also narrows the gap between two descriptor maps and decreases the distinction of descriptors, which can be proved by the decrease of the correct matches. Therefore, the image registration performance peaks at $\lambda=8$ and the setting would be kept in subsequent experiments.

\begin{table}[t]
	\centering
	\setlength{\tabcolsep}{2.0mm}
	\renewcommand\arraystretch{1.5}
	\caption{Impact of $\lambda$ of 1024 keypoints on VIS-IR.}
	\begin{tabular}{l|c|c|c|c|c|c|c}
		\hline
		$\lambda$ & 0.01 & 4    & 8    & 12   & 16   & 20   & 24   \\ \hline
		\# Matches   & 154 &163 &160 &135 &126 &132 &129 \\ \hline
		MS (\%) & 15 &16 &14 &13 &13 &13 &13 \\ \hline
 		RE$_M$   & 4.22 & 3.63 & 2.75 & 2.77 & 3.31 & 2.75 & 3.13 \\ \hline
		\# SR & 36   & 44   & 46   & 44   & 45   & 43   & 44   \\ \hline
	\end{tabular}
\label{tab:tab2}
\end{table}

\textbf{Proposed Components}. We propose three novel modifications: {\small\textcircled{\scriptsize 1}} basic constraints, {\small\textcircled{\scriptsize 2}} recoupled constraints and {\small\textcircled{\scriptsize 3}} new networks for multimodal feature learning. To evaluate the efficiency of our proposals, we choose R2D2, which provides the raw network architecture for losses, as the baseline, and the modifications are successively executed in this framework. As we can see in Table~\ref{tab:tab3}, our basic loss is more suitable for multimodal feature learning and remarkably improves the baseline on all metrics. The recoupled constraints obtain further improvements on feature matching tasks, while the registration performance is comparable to the former. After the new network,~\ie, Super Detector, is equipped, state-of-the-art results are achieved. So far, the proposed components are proved to take positive effects.

\begin{table}[t]
	\centering
	\setlength{\tabcolsep}{1.5mm}
	\renewcommand\arraystretch{1.5}
	\caption{Ablation study of 1024 keypoints on VIS-IR.}
	\begin{tabular}{l|c|c|c|c|c|c}
		\hline
		& \# Corrs & RR(\%) & \# Matches & MS(\%) &RE$_M$ &\# SR\\ \hline
		R2D2 & 213     & 16     & 36        & 4  &3.30 &41  \\ \hline
		+{\textcircled{\scriptsize 1}}  & 307     & 30     & 83    & 8   & 2.81 &45   \\ \hline
		+{\textcircled{\scriptsize 1}}+{\textcircled{\scriptsize 2}}   & 346     & 33     & 112       & 11  &2.93 &46   \\ \hline
		+{\textcircled{\scriptsize 1}}+{\textcircled{\scriptsize 2}}+{\textcircled{\scriptsize 3}}    & 415     & 40     & 160       & 15  &2.75 &46   \\ \hline
	\end{tabular}
	\label{tab:tab3}
\end{table}

To gain an insight into the impacts of our proposals, we visualize the detection score maps, what we discuss throughout our formulation, under different configurations in Fig.~\ref{fig:abla}. As shown in the second and the third columns of images, while the local peaking loss guides R2D2 to generate discrete detection score, it leads to bulks of detection basic constraints. It can be explained that $\lambda=8$ moderates the impact of local peaking in basic constraints. These lumped detected features intend to repeat and be matched within an acceptable error so that the matching performance is improved. After recoupling the constraints, the edge-based priori makes the detection gather in areas with rich textures, which is expected to obtain further improvements. Finally, the Super Detector equipped with learnable local non-maximum suppression blocks introduces a strong inductive bias to discretize the detection score. The discrete detection score must tighten weighted description loss and repeatability loss, which is believed to help the joint learning and improve the accuracy of keypoint location.

\renewcommand\thefigure{7}
\begin{figure*}[b]
	\centering
	\includegraphics[width = 2.8cm]{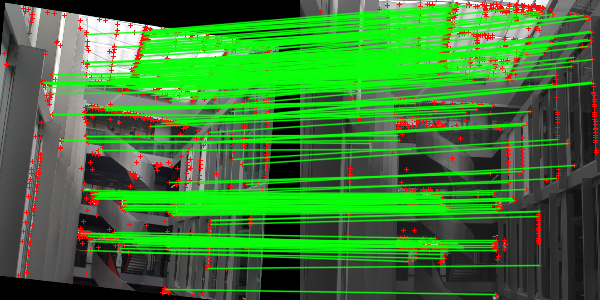} \hspace{-0.6em}
	\includegraphics[width = 2.8cm]{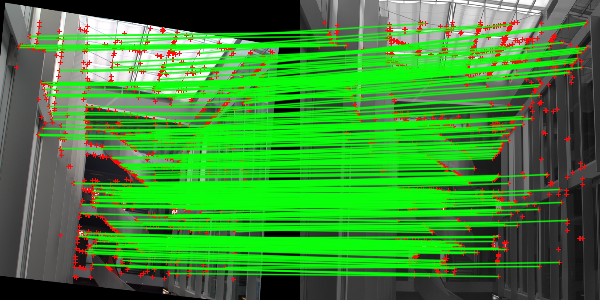} \hspace{-0.6em}
	\includegraphics[width = 2.8cm]{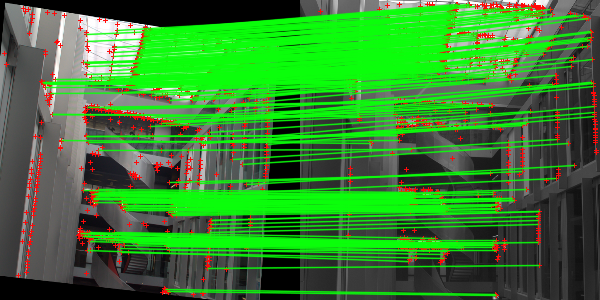} \hspace{-0.6em}
	\includegraphics[width = 2.8cm]{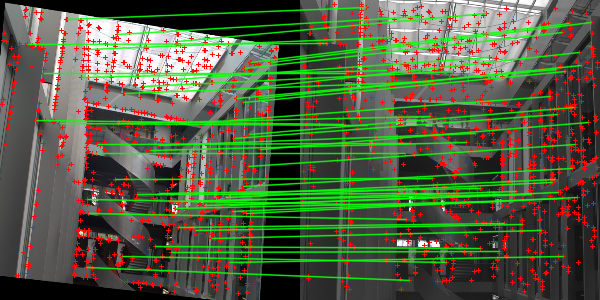} \hspace{-0.6em}
	\includegraphics[width = 2.8cm]{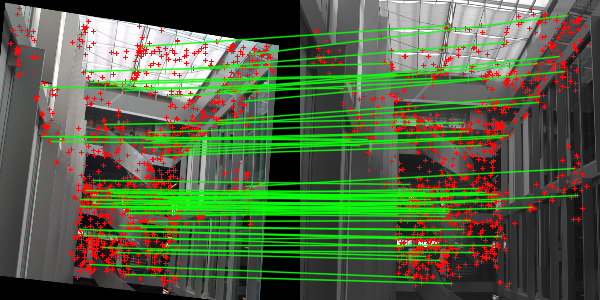} \hspace{-0.6em}
	\includegraphics[width = 2.8cm]{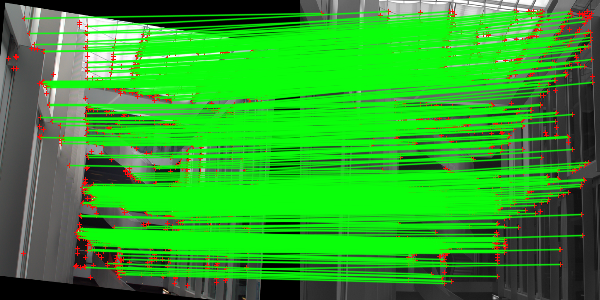}\\
	\vspace{0.12em}
	\includegraphics[width = 2.8cm]{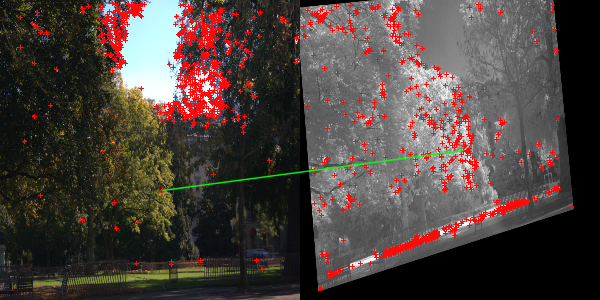} \hspace{-0.6em}
	\includegraphics[width = 2.8cm]{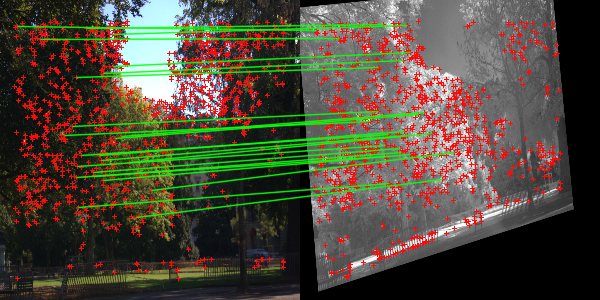} \hspace{-0.6em}
	\includegraphics[width = 2.8cm]{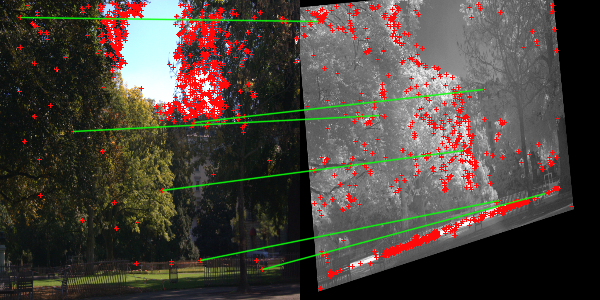} \hspace{-0.6em}
	\includegraphics[width = 2.8cm]{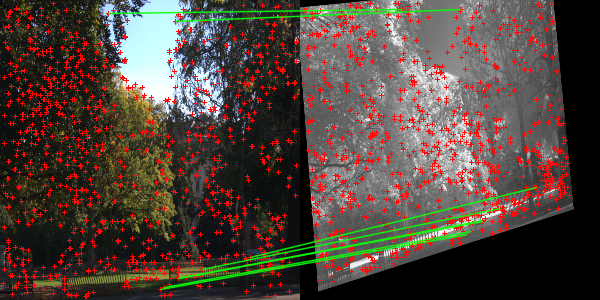} \hspace{-0.6em}
	\includegraphics[width = 2.8cm]{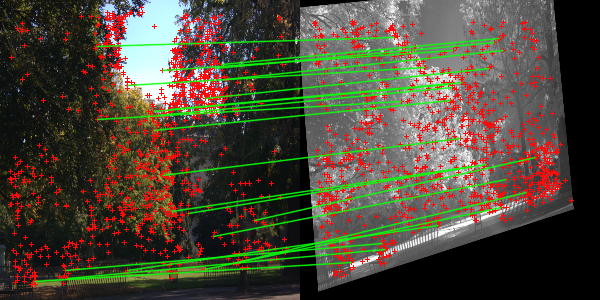} \hspace{-0.6em}
	\includegraphics[width = 2.8cm]{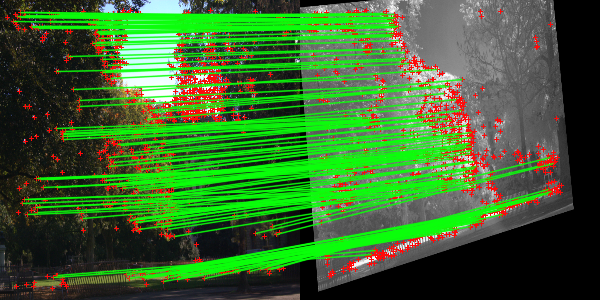}\\
	\vspace{0.12em}
	\includegraphics[width = 2.8cm]{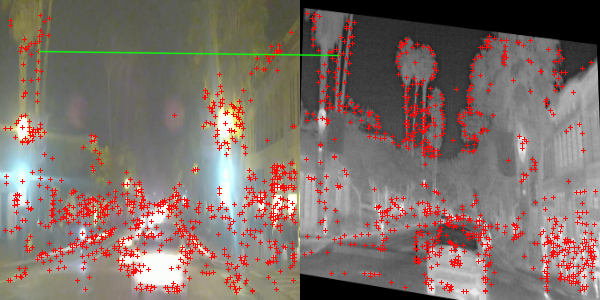} \hspace{-0.6em}
	\includegraphics[width = 2.8cm]{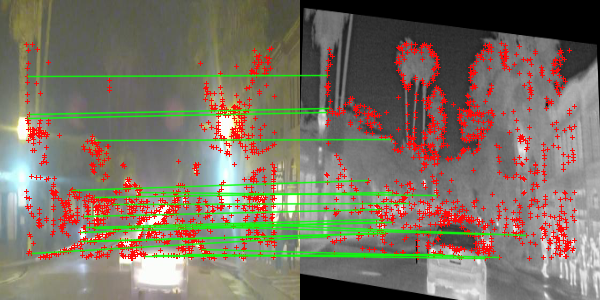} \hspace{-0.6em}
	\includegraphics[width = 2.8cm]{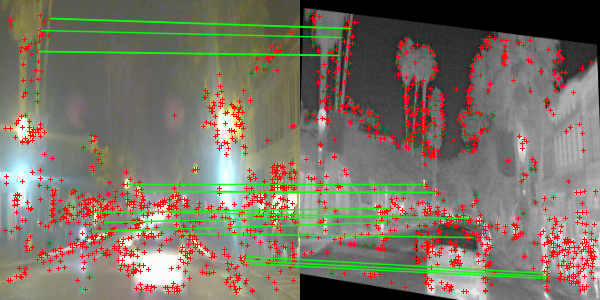} \hspace{-0.6em}
	\includegraphics[width = 2.8cm]{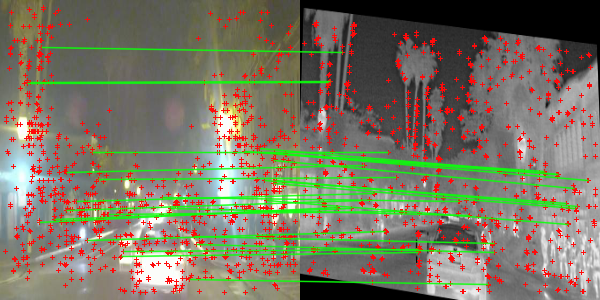} \hspace{-0.6em}
	\includegraphics[width = 2.8cm]{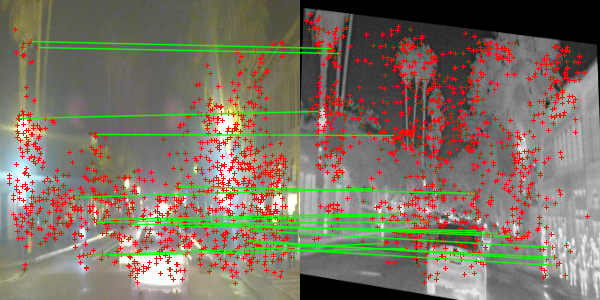} \hspace{-0.6em}
	\includegraphics[width = 2.8cm]{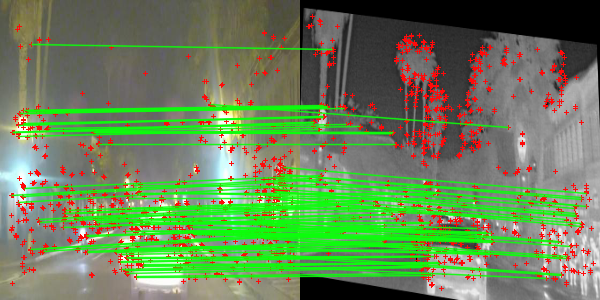}\\
	\vspace{0.12em}
	\includegraphics[width = 2.8cm]{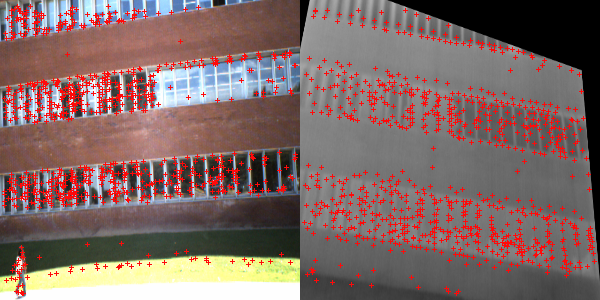} \hspace{-0.6em}
	\includegraphics[width = 2.8cm]{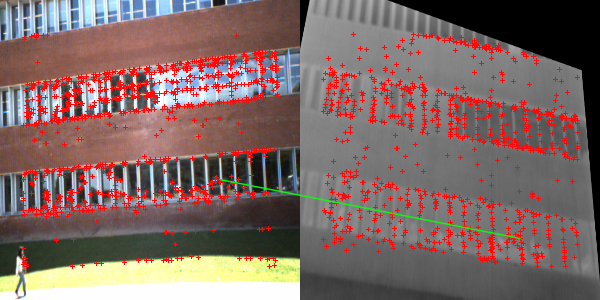} \hspace{-0.6em}
	\includegraphics[width = 2.8cm]{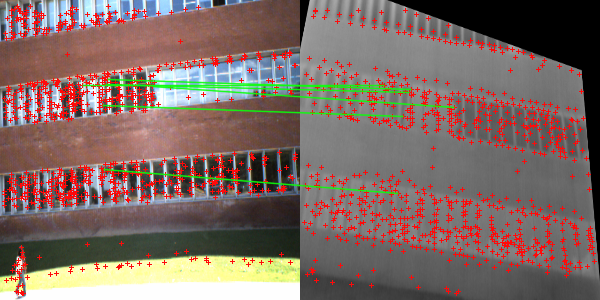} \hspace{-0.6em}
	\includegraphics[width = 2.8cm]{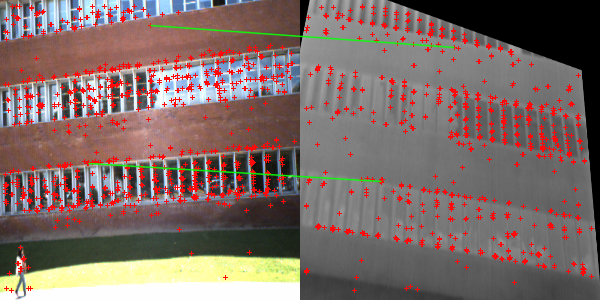} \hspace{-0.6em}
	\includegraphics[width = 2.8cm]{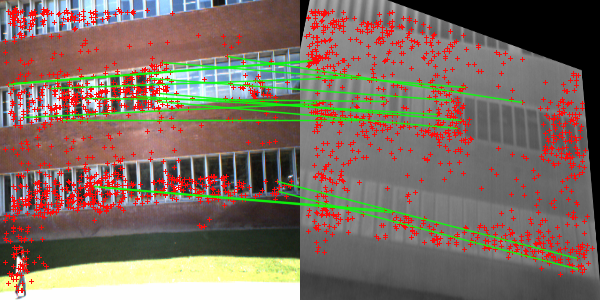} \hspace{-0.6em}
	\includegraphics[width = 2.8cm]{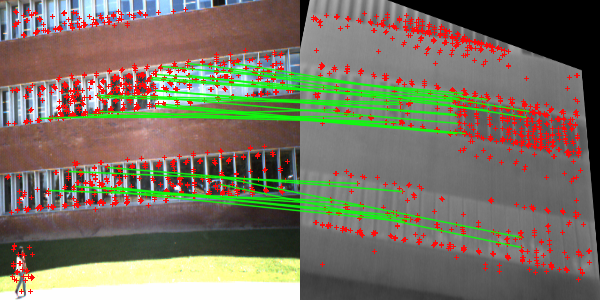}\\
	\vspace{0.12em}
	\includegraphics[width = 2.8cm]{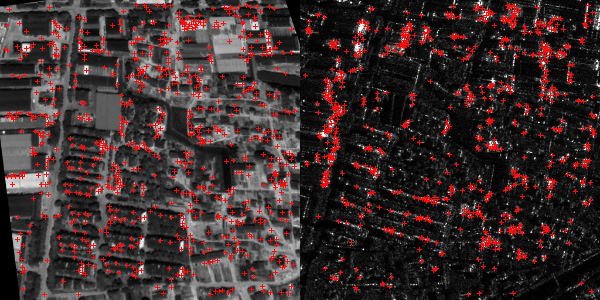} \hspace{-0.6em}
	\includegraphics[width = 2.8cm]{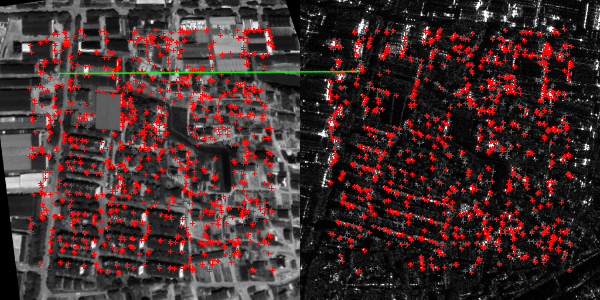} \hspace{-0.6em}
	\includegraphics[width = 2.8cm]{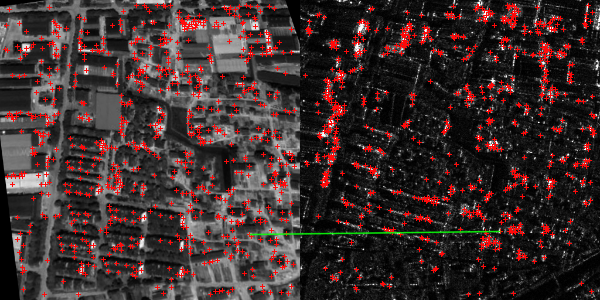} \hspace{-0.6em}
	\includegraphics[width = 2.8cm]{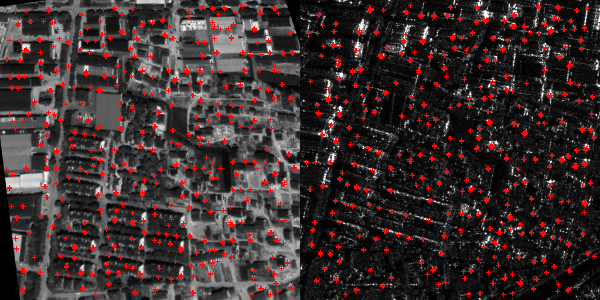} \hspace{-0.6em}
	\includegraphics[width = 2.8cm]{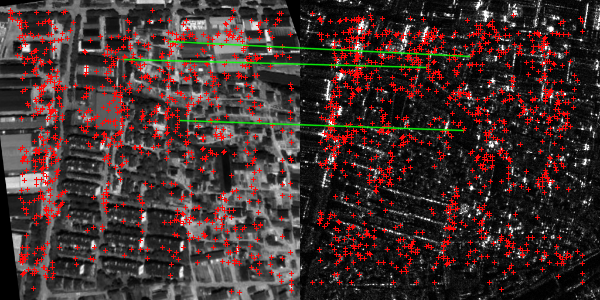} \hspace{-0.6em}
	\includegraphics[width = 2.8cm]{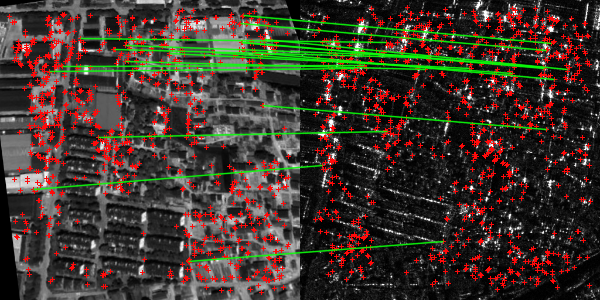}\\
	\vspace{0.12em}
	\includegraphics[width = 2.8cm]{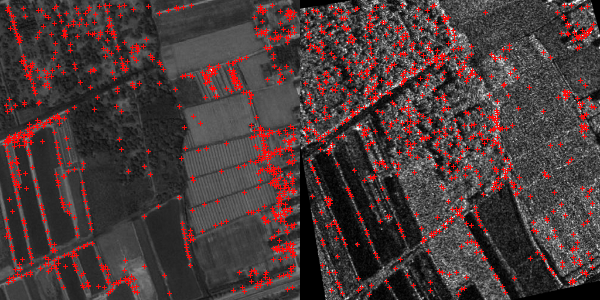} \hspace{-0.6em}
	\includegraphics[width = 2.8cm]{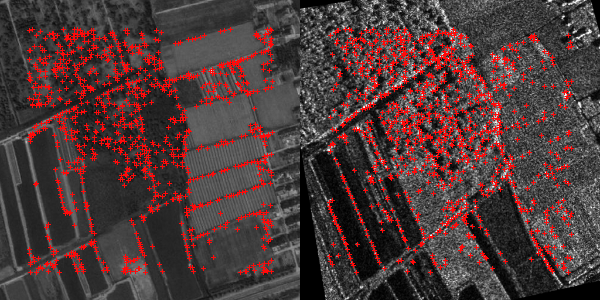} \hspace{-0.6em}
	\includegraphics[width = 2.8cm]{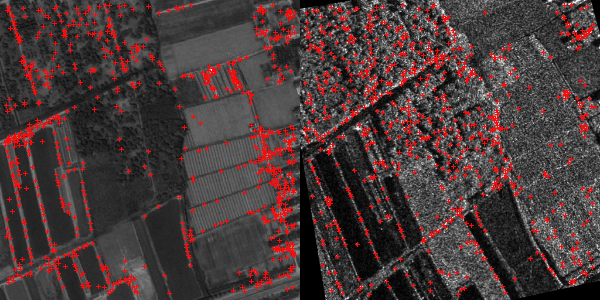} \hspace{-0.6em}
	\includegraphics[width = 2.8cm]{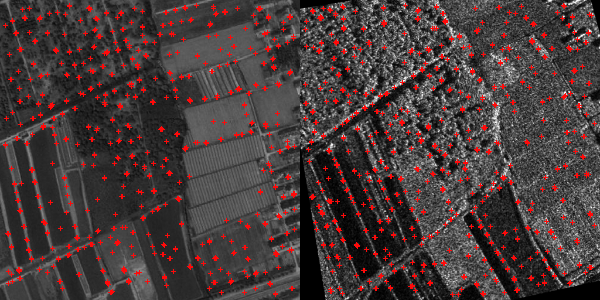} \hspace{-0.6em}
	\includegraphics[width = 2.8cm]{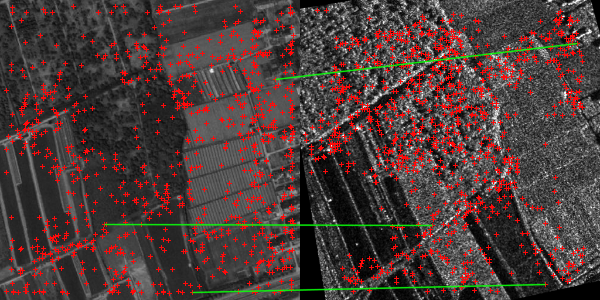} \hspace{-0.6em}
	\includegraphics[width = 2.8cm]{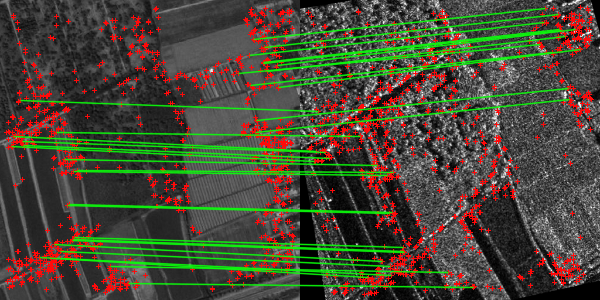}\\
	\raggedright
	{\footnotesize \hspace{1.4cm} SIFT\hspace{2.2cm} RIFT\hspace{1.90cm} DOG+HN\hspace{1.95cm} R2D2\hspace{1.95cm} CMMNet\hspace{1.80cm} ReDFeat}
	\caption{Visualization of matching performance. 1024 keypoints are extracted by different algorithms and marked in red `+'. Descriptors are matched by the bidirectional nearest neighbor matching. Validated at a threshold of 3px, the correct matches are linked by the green lines.}
	\label{fig:mp_viz}
\end{figure*}

\renewcommand\thefigure{5}
\begin{figure}
	\centering
	\includegraphics[width=1.0\linewidth]{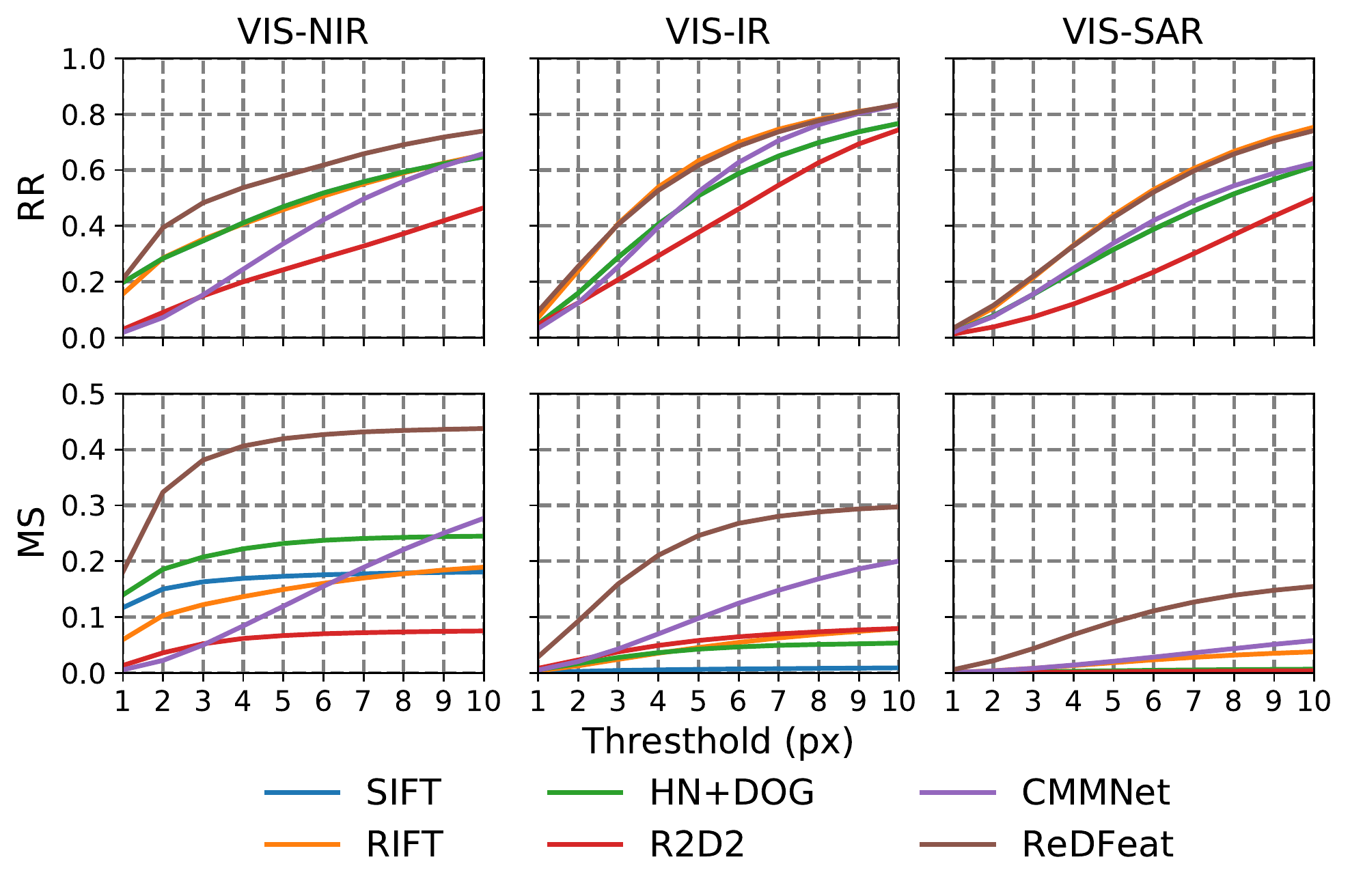}
	\caption{Feature matching performance of 1024 keypoints of the state-of-the-art methods in our benchmark. Repeatable rate (RR) and matching score (MS) are computed at thresholds up to 10 pixels are drawn in curves.}
	\label{fig:matchthres}
\end{figure}

\renewcommand\thefigure{6}
\begin{figure}[!ht]
	\centering
	\includegraphics[width=0.95\linewidth]{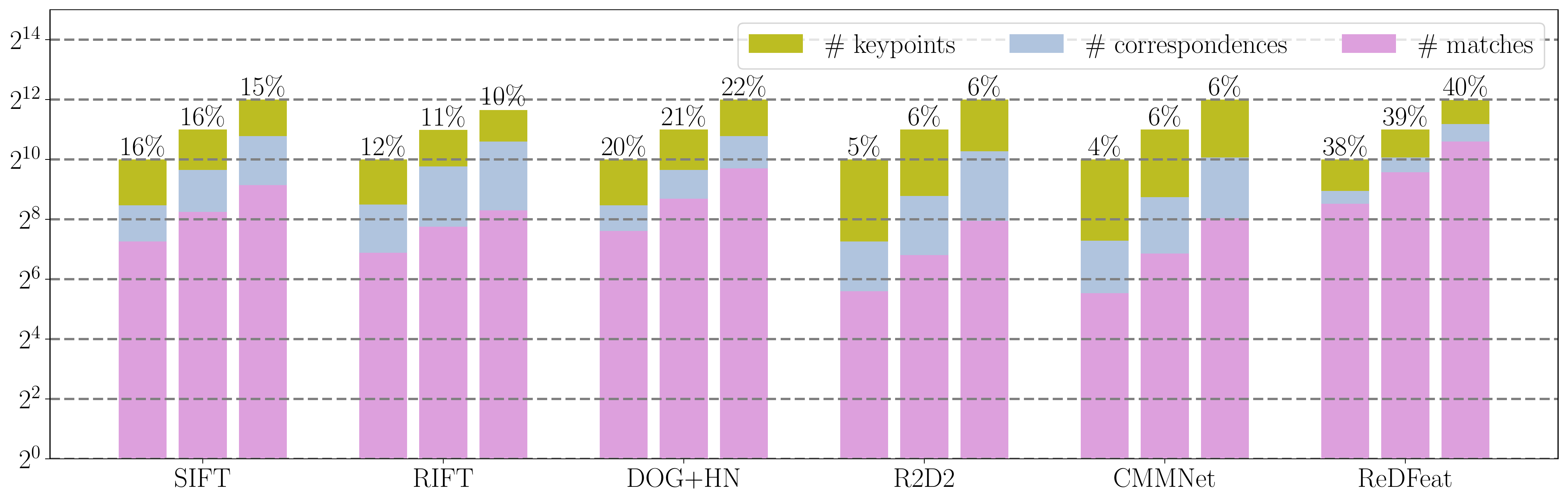}\\
	\footnotesize{(a) Matching Performance on VIS-NIR}\\
	\includegraphics[width=0.95\linewidth]{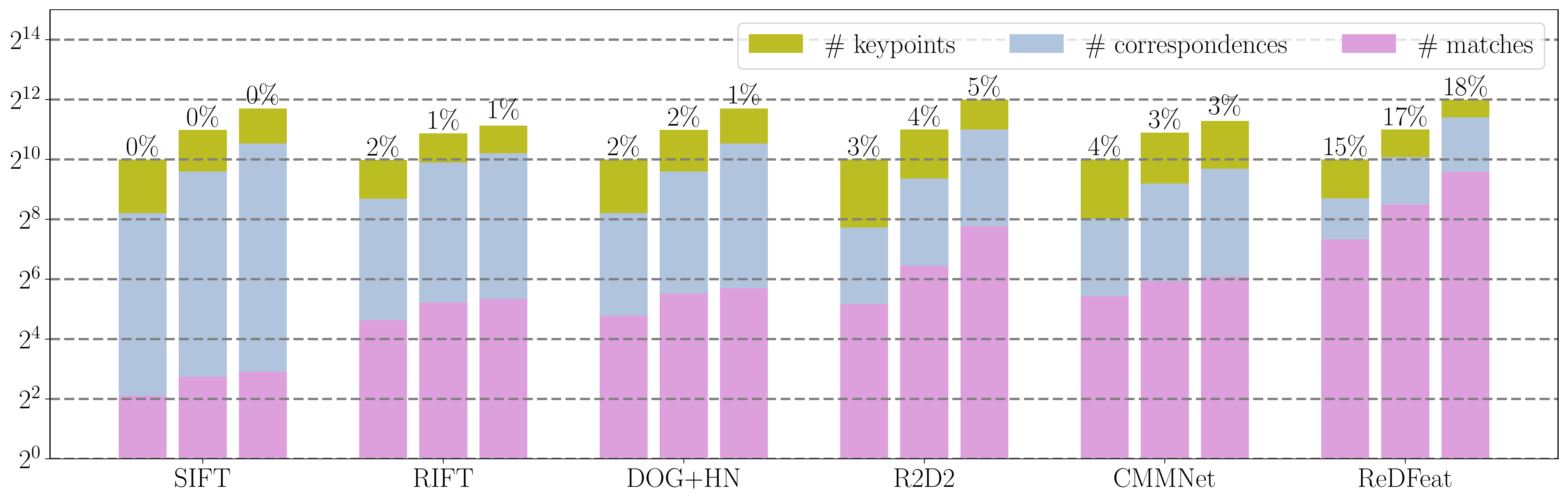}\\
	\footnotesize{(b) Matching Performance on VIS-IR}\\
	\includegraphics[width=0.95\linewidth]{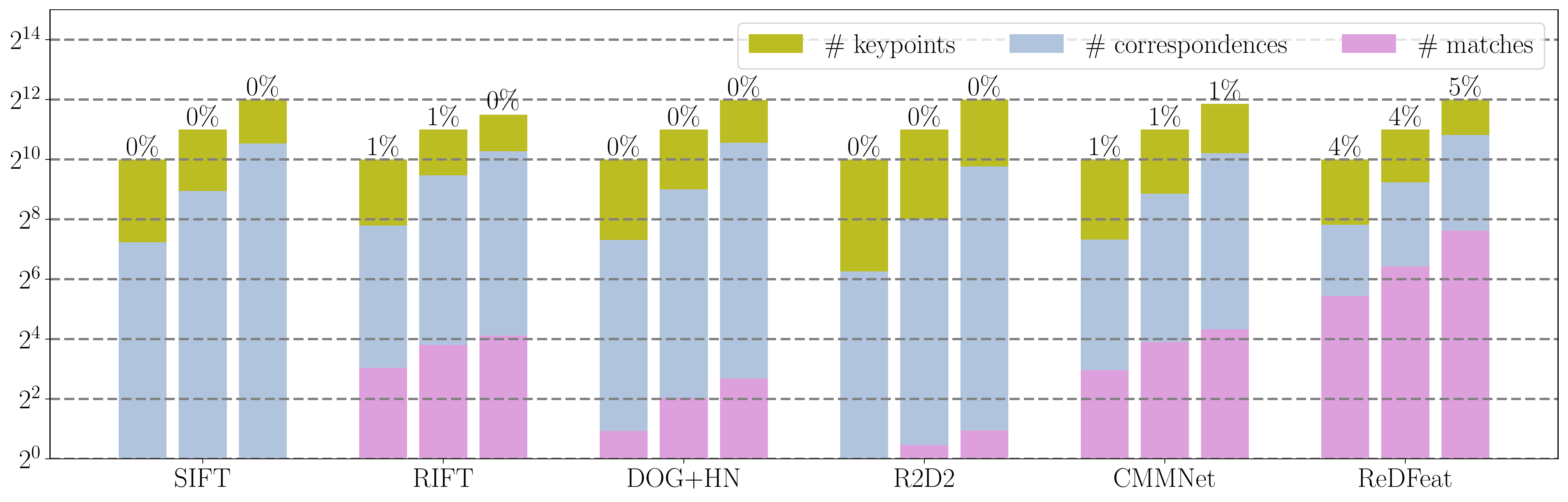}\\
	\footnotesize{(c) Matching Performances on VIS-SAR}\\
	\caption{Feature matching performance of 1024, 2048 and 4096 keypoints at 3px. The numbers of extracted keypoints, correspondences and correct matches are drawn in bars with different colors. Matching scores of three numbers of keypoints are shown on the bars.}	
	\label{fig:mp}
\end{figure}

\subsection{Feature Matching Performance}
The feature matching performance of 1024 keypoints of SIFT, RIFT, HN, R2D2, CMMNet, and our RedFeat on three subsets is quantified in Fig.~\ref{fig:matchthres}, in which RR and MS are selected as the primary metrics and calculated at varying thresholds. As we can see, we achieve the state-of-the-art RR and MS at all thresholds on three subsets, which demonstrates that we obtain more robust descriptors while detecting more precise and repeatable keypoints. As for MS, we gain large margins on all subsets at varying thresholds. Especially on the most challenging subset, VIS-SAR, our MS seems to be several times higher than the second place CMMNet. It is worth mentioning that R2D2 employing pre-trained models for initialization still fails to optimize the description on VIS-SAR, which confirms the significance of our recoupling strategy.

More quantitative performance of 1024, 2048 and 4096 keypoints at a threshold of 3px is shown in Fig.~\ref{fig:mp}. The matching score that is key index for feature matching performance is deliberately highlighted on the bars. Except for the number of correspondences on VIS-IR and VIS-SAR, our ReDFeat achieve the best scores on all metrics at 3px. Note that, the handcrafted local-maximums-searching-based detector might fail to extract large numbers of keypoints for some image pairs, which demonstrates the superiority and the flexibility of the learnable detection.

Qualitative performance is shown in Fig.~\ref{fig:mp_viz}. Compared to R2D2 and CMMNet, our detected points seem to be more rationally distributed in the textured area,~\ie, intensely but not too intensely. However, traditional detectors employed by SIFT, RIFT, and DOG-HN seemingly generate more interpretable results that strictly attach to the edges or corners. Especially, RIFT detects scattered corner points, which are proved by RR shown in Fig.~\ref{fig:matchthres} and the number of correspondences shown in Fig.~\ref{fig:mp}, in not so salient regions. The weakness of the deep learning detector can be attributed to the flaws of the training set, which cannot provide strict correspondences so that the keypoints are not precisely located. Despite the advantages of traditional detectors, the deep learning one-stage and two-stage methods show the superiority of deep learning on the feature description. In our method that the description and detection are mutually guided in our recoupling strategy, the hard descriptors are better optimized, which makes significant progress in matching performance.

\subsection{Image Registration Performance}
Successful registration rates of 1024 keypoints of those algorithms are drawn in Fig.~\ref{fig:reprojthres}. Note that, we use two measures of projection error (RE) on the three subsets according to the quality of the ground truths. Nevertheless, our ReDFeat obtains more successfully registered images pairs in each case. And the margin is more prominent on VIS-SAR that is the most challenging. Moreover, the weak performance of CMMNet on VIS-NIR highlights the important of keypoint location,~\ie, the registration performance depends on MS at low thresholds. The problem is well tackled by recoupled constraints and Super Detector in our method, as proved in ablation study.

\renewcommand\thefigure{8}
\begin{figure}
	\centering
	\includegraphics[width=1.0\linewidth]{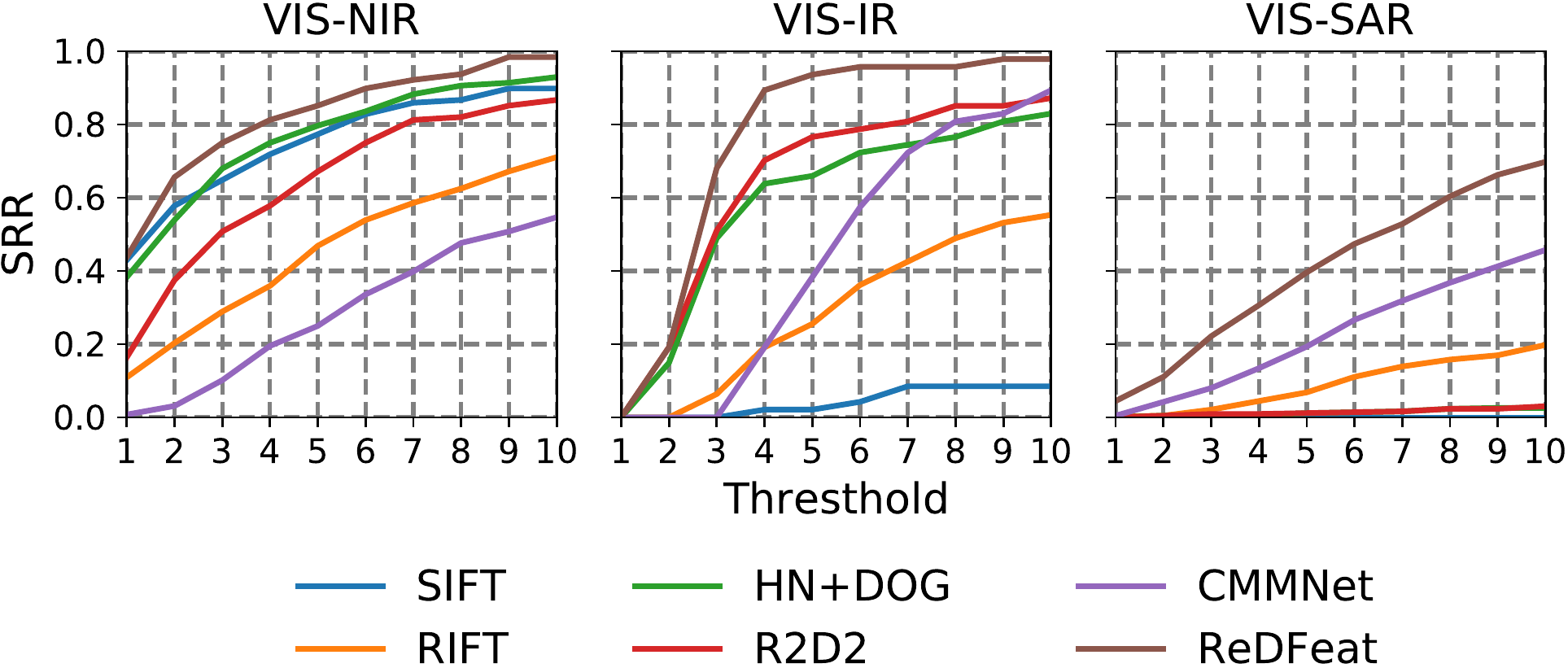}
	\caption{Successful registration rate (SSR) of $1024$ keypoints at varying thresholds up to 10. Note that, different measures of reprojection error,~\ie, the meanings of the threshold are used on VIS-IR. }
	\label{fig:reprojthres}
\end{figure}

The distributions of reprojection errors of 1024, 2048 and 4096 keypoints are illustrated in Fig.~\ref{fig:ir}.
Except on the VIS-NIR with $2048$ and $4096$ keypoints, our method achieves the most SR and the lowest mean RE. Particularly, while greatly boosting SR on VIS-SAR, our method gets the mostly precise image registration performance. As for the tiny disparity of RE among SIFT, DOG-HN, and ReDFeat on VIS-NIR, it can be explained by the small discrepancy between visible and near-infrared images and the accuracy of handcrafted keypoint location as mentioned above.

Some examples, in which only our ReDFeat succeeds, are shown in Fig.~\ref{fig:ir_viz}. Although RIFT, R2D2 and CMMNet estimate approximate transforms in some cases from VIS-NIR and VIS-IR, the accuracy of registration does not meet the expectations. On samples of VIS-SAR, the other alternatives even fail to receive a rough result, which is consistent with the feature matching performance. Generally, with the help of recoupled constraints and Super Detector, our method can learn robust cross-modal features that indeed boost the performance of cross-modal image registration.

\renewcommand\thefigure{9}
\begin{figure}[t]
	\centering
	\includegraphics[width=0.9\linewidth]{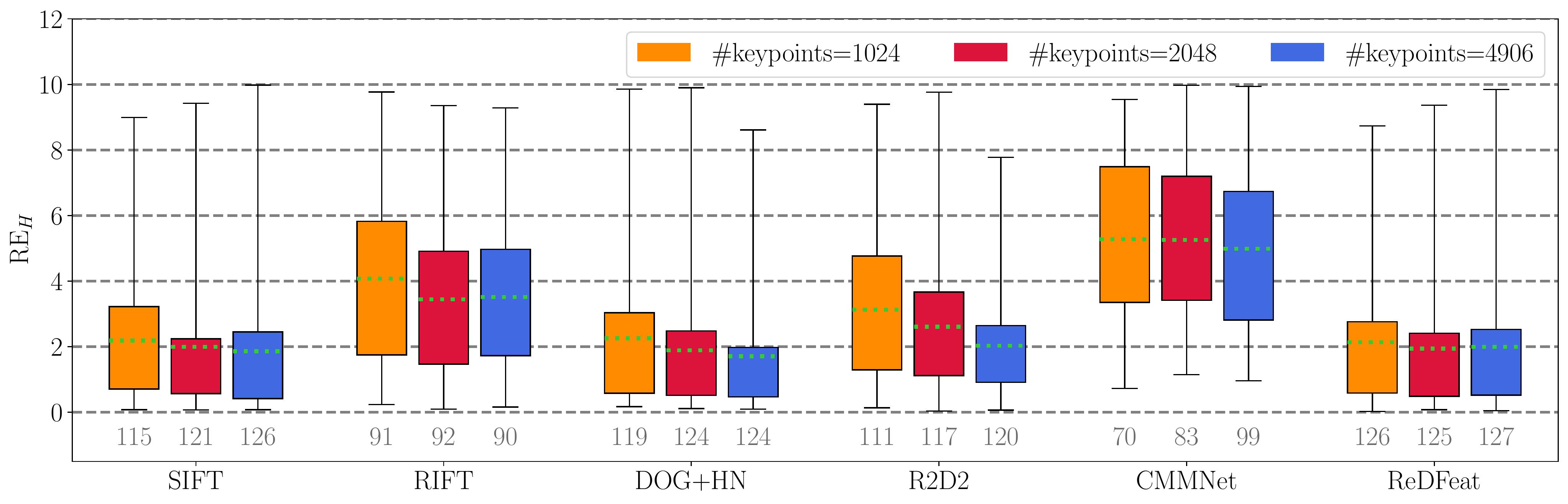}\\
	\footnotesize{(a) Reprojection Errors on VIS-NIR}\\
	\vspace{0.05in}
	\includegraphics[width=0.9\linewidth]{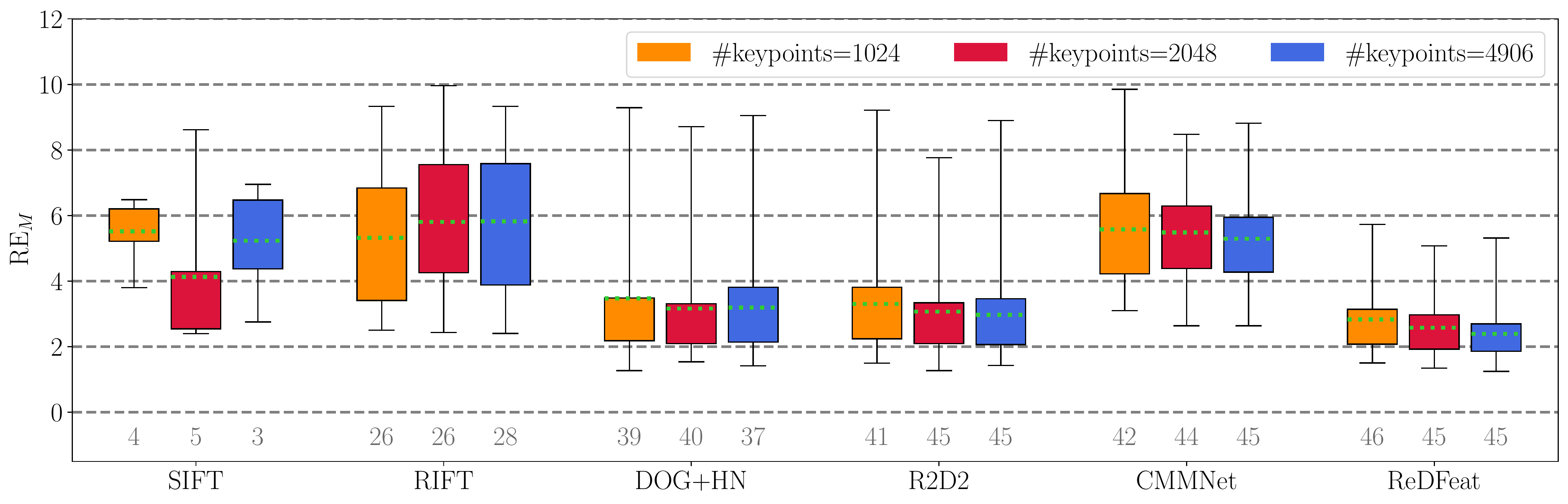}\\
	\footnotesize{(b) Reprojection Errors on VIS-IR}\\
	\vspace{0.05in}
	\includegraphics[width=0.9\linewidth]{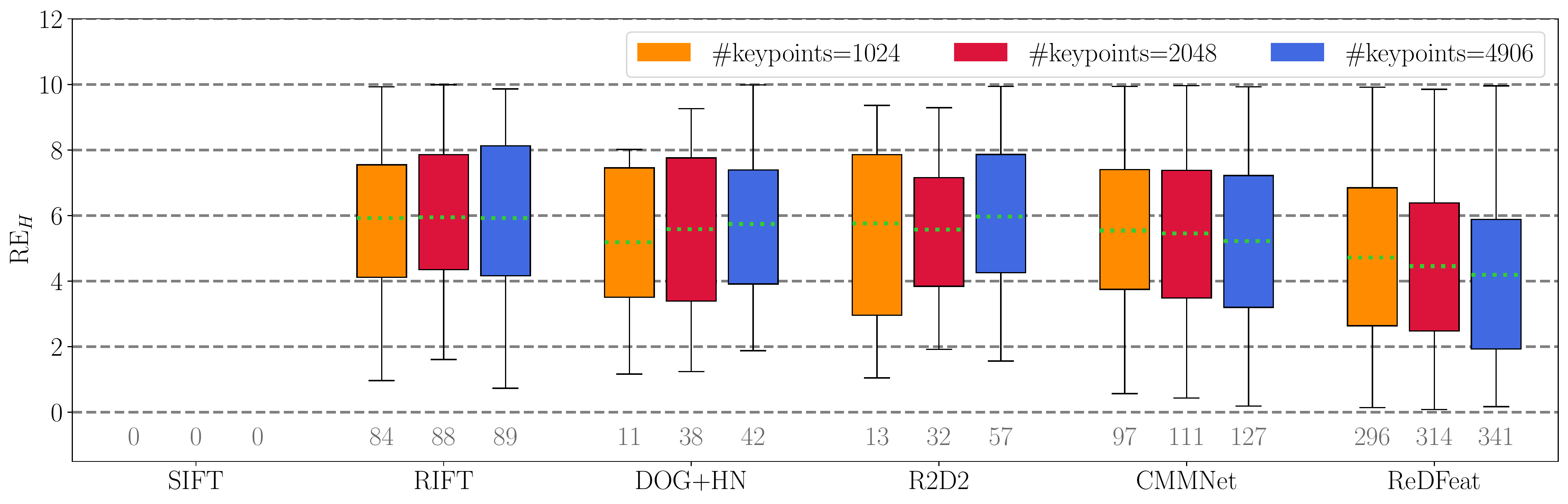}\\
	\footnotesize{(c) Reprojection Errors on VIS-SAR}\\
	\caption{Reprojection errors of 1024, 2048 and 4096 keypoints of the state-of-the-art methods in our benchmark. Different numbers of keypoints are extracted and drawn in different colors. The distribution of the projection errors of the successfully registered images (SR) at 10 are drawn in box plots, in which the green dash lines indicate the mean of data; the boxes cover samples from the $25$th to $75$th percentile of the errors; the maximums or minimums are marked by caps. The numbers of SR at $10$ are shown under corresponding box plots. }	
	\label{fig:ir}
\end{figure}

\begin{table}[t]
	\setlength{\tabcolsep}{1.5mm}
	\renewcommand\arraystretch{1.5}
	\centering
	\caption{Average runtime of different methods in our benchmark. The average sizes of images are given, and the runtime is counted in millisecond (ms).}
	\begin{tabular}{l|c|c|c|c|c|c}
		\hline
		Time (ms) & SIFT & RIFT & DOG+HN & R2D2 & CMMNet & ReDFeat \\ \hline
		
		\begin{tabular}[c]{@{}l@{}}VIS-NIR\\ (971$\times$682)\end{tabular}
		& 236  & 3186 & 351    & 59   & 1790   & 94      \\ \hline
		\begin{tabular}[c]{@{}l@{}}VIS-IR\\ (528$\times$320)\end{tabular}
		& 68   & 1984 & 229    & 54   & 458    & 74      \\ \hline
		\begin{tabular}[c]{@{}l@{}}VIS-SAR\\ (512$\times$512)\end{tabular}
		  & 97   & 2530 & 263    & 56   & 675    & 85      \\ \hline
	\end{tabular}
	\label{tab:rt}
\end{table}

\renewcommand\thefigure{10}
\begin{figure*}[t]
	\centering
	\includegraphics[width = 2.9cm]{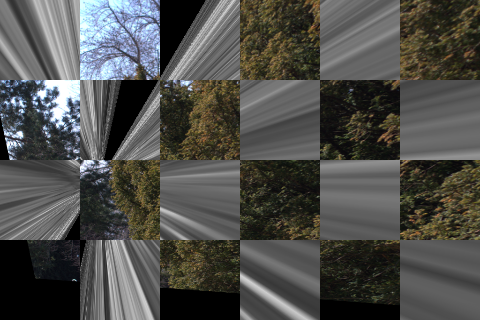} \hspace{-0.6em}
	\includegraphics[width = 2.9cm]{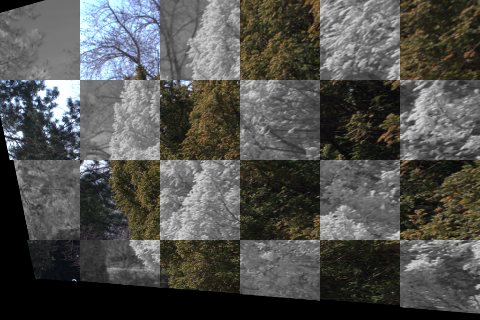} \hspace{-0.6em}
	\includegraphics[width = 2.9cm]{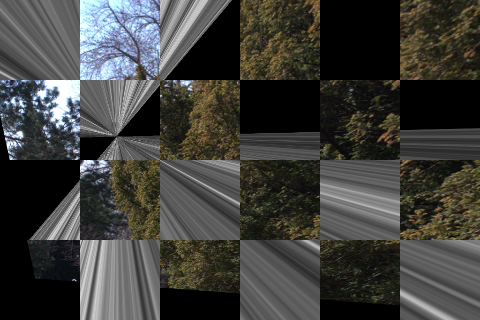} \hspace{-0.6em}
	\includegraphics[width = 2.9cm]{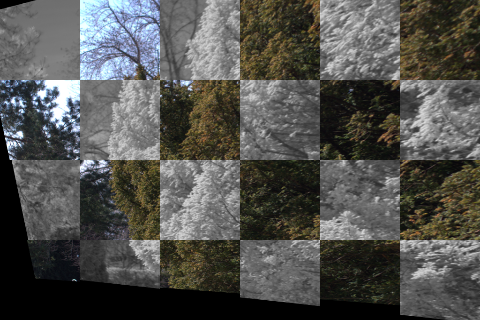} \hspace{-0.6em}
	\includegraphics[width = 2.9cm]{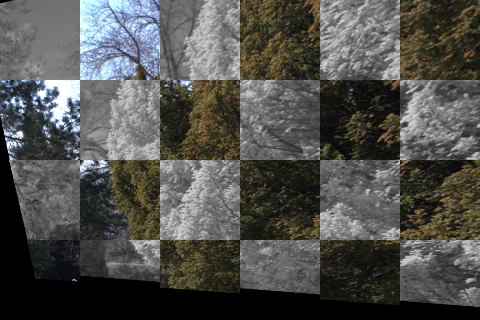} \hspace{-0.6em}
	\includegraphics[width = 2.9cm]{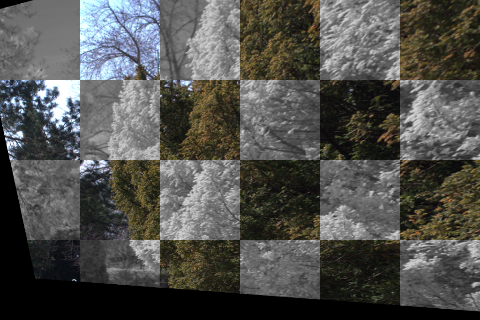}\\
	\vspace{0.12em}
	\includegraphics[width = 2.9cm]{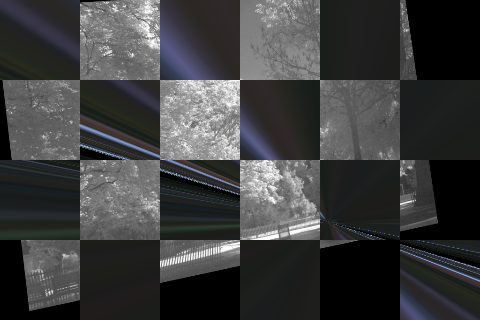} \hspace{-0.6em}
	\includegraphics[width = 2.9cm]{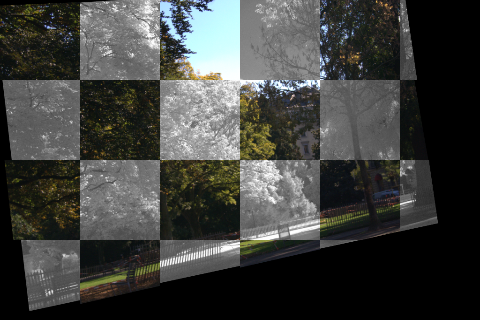} \hspace{-0.6em}
	\includegraphics[width = 2.9cm]{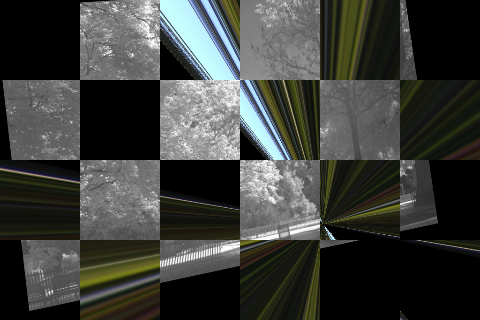} \hspace{-0.6em}
	\includegraphics[width = 2.9cm]{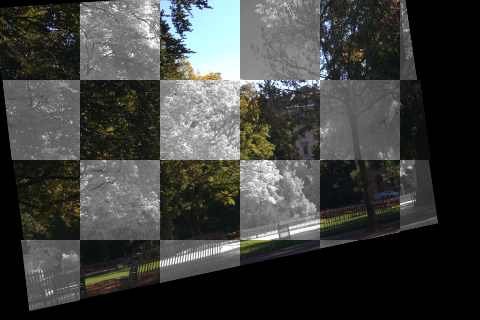} \hspace{-0.6em}
	\includegraphics[width = 2.9cm]{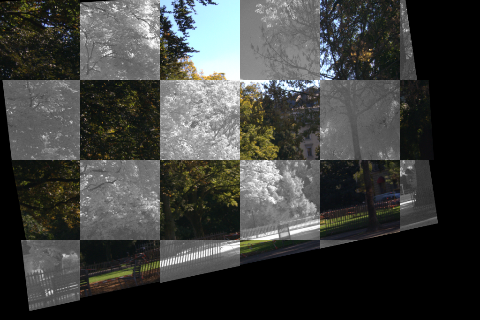} \hspace{-0.6em}
	\includegraphics[width = 2.9cm]{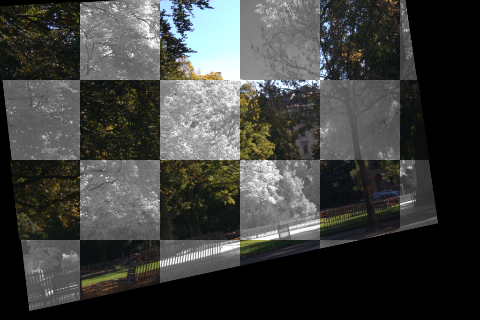}\\
	\vspace{0.12em}
	\includegraphics[width = 2.9cm]{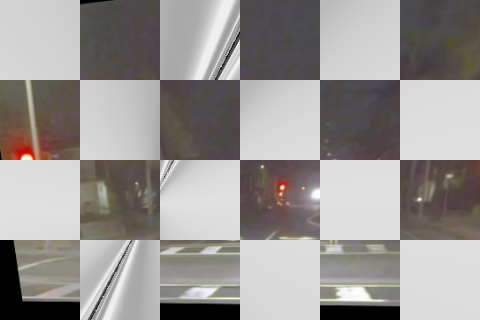} \hspace{-0.6em}
	\includegraphics[width = 2.9cm]{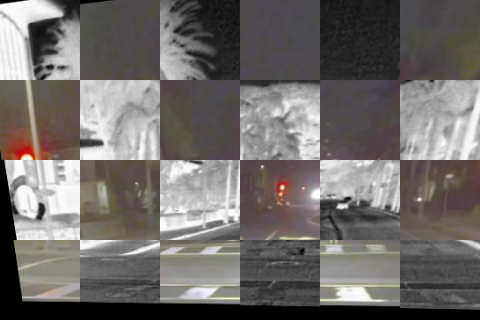} \hspace{-0.6em}
	\includegraphics[width = 2.9cm]{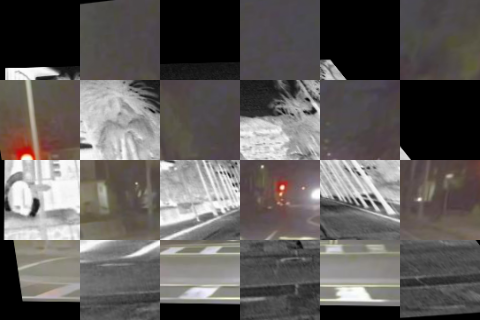} \hspace{-0.6em}
	\includegraphics[width = 2.9cm]{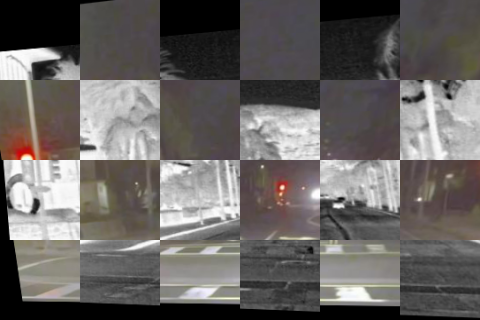} \hspace{-0.6em}
	\includegraphics[width = 2.9cm]{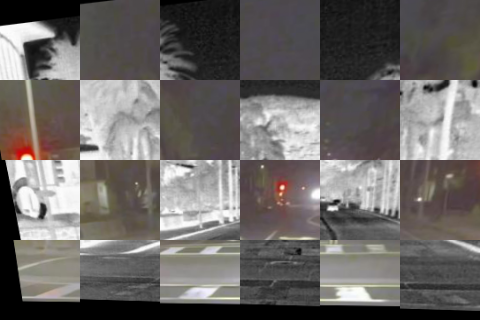} \hspace{-0.6em}
	\includegraphics[width = 2.9cm]{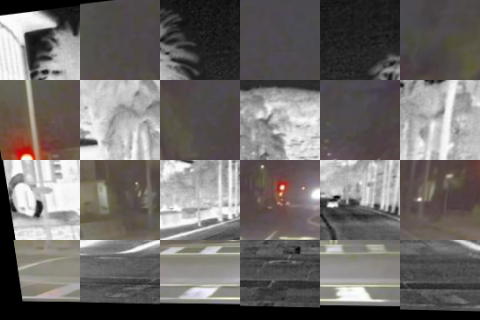}\\
	\vspace{0.12em}
	\includegraphics[width = 2.9cm]{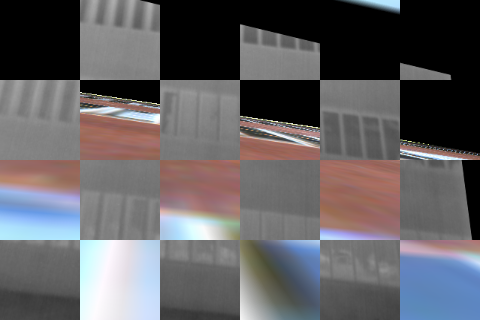} \hspace{-0.6em}
	\includegraphics[width = 2.9cm]{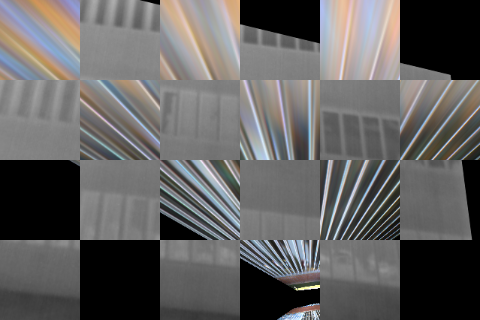} \hspace{-0.6em}
	\includegraphics[width = 2.9cm]{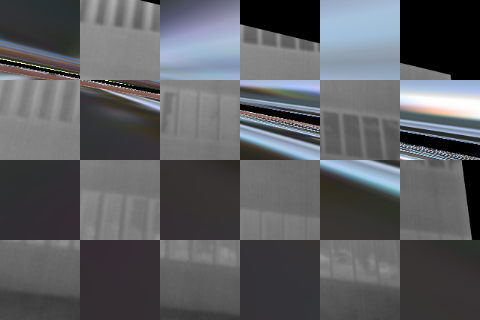} \hspace{-0.6em}
	\includegraphics[width = 2.9cm]{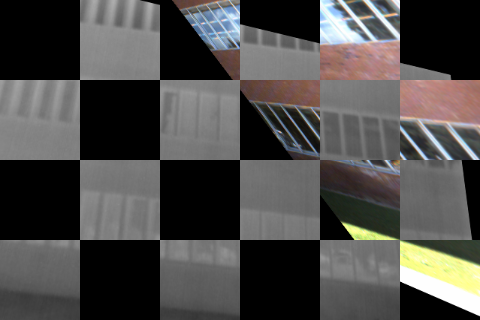} \hspace{-0.6em}
	\includegraphics[width = 2.9cm]{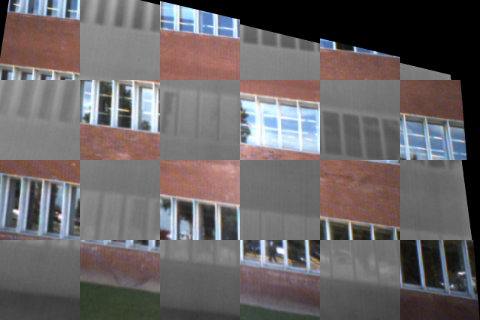} \hspace{-0.6em}
	\includegraphics[width = 2.9cm]{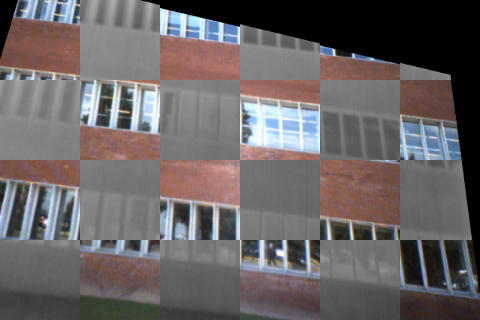}\\
	\vspace{0.12em}
	\includegraphics[width = 2.9cm]{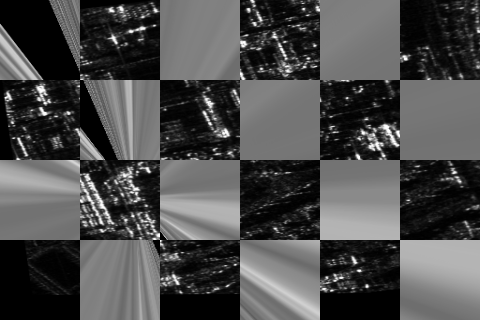} \hspace{-0.6em}
	\includegraphics[width = 2.9cm]{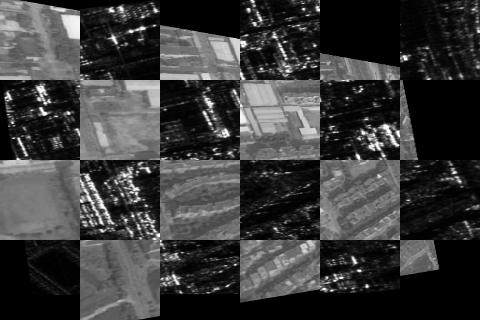} \hspace{-0.6em}
	\includegraphics[width = 2.9cm]{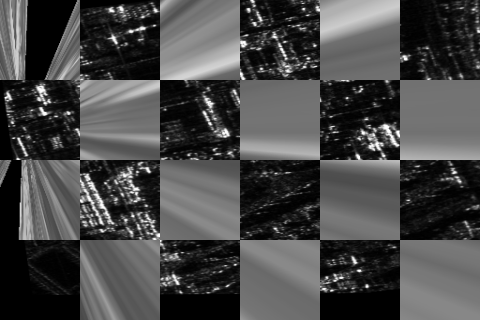} \hspace{-0.6em}
	\includegraphics[width = 2.9cm]{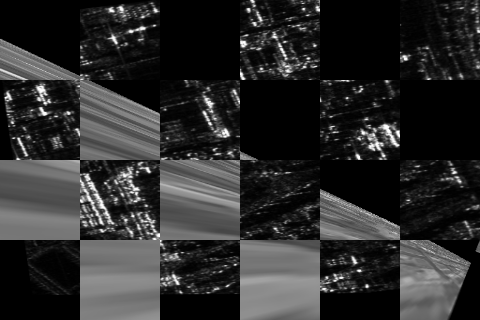} \hspace{-0.6em}
	\includegraphics[width = 2.9cm]{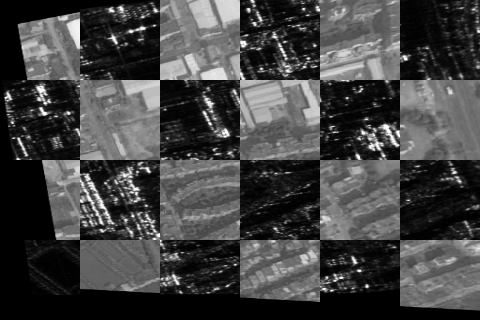} \hspace{-0.6em}
	\includegraphics[width = 2.9cm]{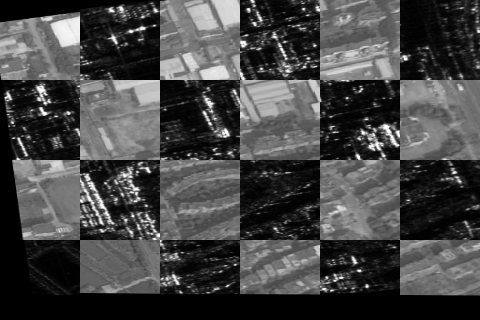}\\
	\vspace{0.12em}
	\includegraphics[width = 2.9cm]{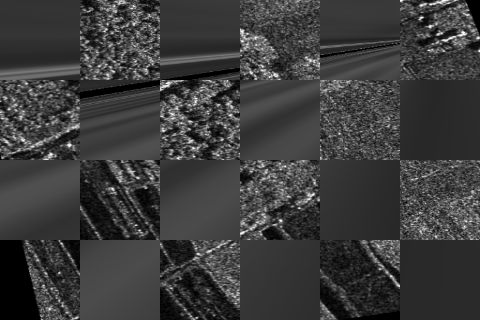} \hspace{-0.6em}
	\includegraphics[width = 2.9cm]{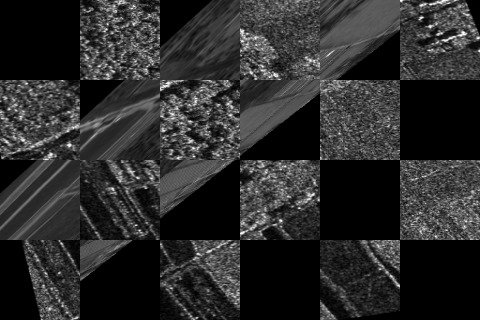} \hspace{-0.6em}
	\includegraphics[width = 2.9cm]{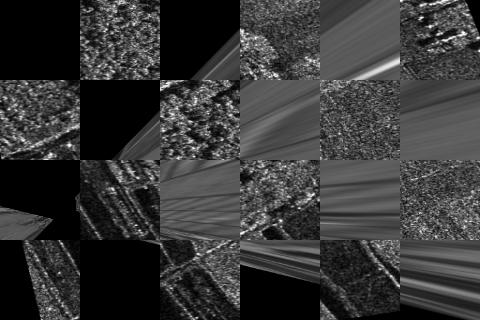} \hspace{-0.6em}
	\includegraphics[width = 2.9cm]{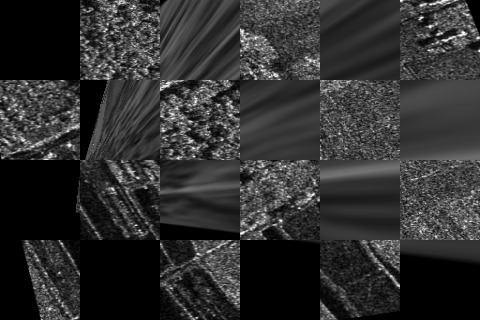} \hspace{-0.6em}
	\includegraphics[width = 2.9cm]{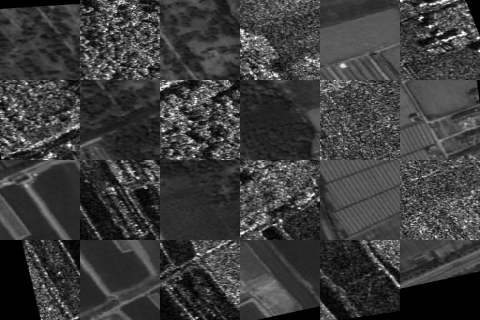} \hspace{-0.6em}
	\includegraphics[width = 2.9cm]{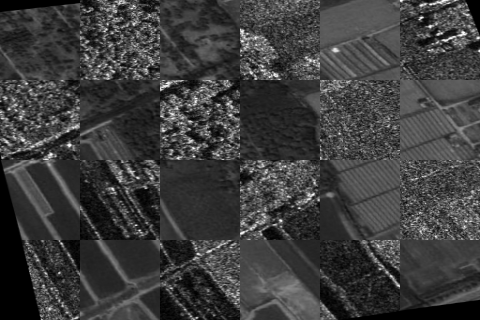}\\
	\raggedright
	{\footnotesize \hspace{1.4cm} SIFT\hspace{2.2cm} RIFT\hspace{1.90cm} DOG+HN\hspace{1.95cm} R2D2\hspace{1.95cm} CMMNet\hspace{1.80cm} ReDFeat}
	\caption{Visualization of image registration performance. 1024 features are extracted by different algorithms and matched by bidirectional nearest neighbor matching. Note that, the images in 2nd, 4th and 6th rows are corresponding to the images in Fig.~\ref{fig:mp_viz}.}
	\label{fig:ir_viz}
\end{figure*}

\subsection{Runtime}
Time consumption is important for feature extraction. Because SIFT, RIFT, DOG+HN and CMMNet employ handcrafted detectors, their computation complexities are hard to calculate. And we just report the average runtime in three test sets in Table~\ref{tab:rt}. All of them are implemented in Python, except RIFT which is implemented in Matlab. All methods are run on an Intel Xeon Silver 4210R CPU and an NVIDIA RTX3090 GPU. As we can see, R2D2 consumes the least time to extract features for each image. And benefiting from the parallel computation on GPU, its runtime is not sensitive to the image size. Because of the complex operations in Super Detector, ReDFeat takes more time to finish the extraction. However, the improvements of our method are believed to be worth the increased runtime. All methods seem to be time-consuming except SIFT which takes the same order of magnitude of time as R2D2 and ReDFeat. Generally, we significantly improve the performance of features with few extra costs.

\section{Conclusion}
In this paper, we take the ill-posed detection in joint detection and description framework as the start point, and propose the recoupled constraints for multimodal feature learning. Firstly, based on the efforts from related works, we reformulate the repeatability loss and the local peaking loss for detection, as well as the contrastive loss for description in multimodal scenario. Then, to recouple the constraints of the detection and description, we propose the mutual weighting strategy, in which the robust features are forced to achieve desired detected probabilities that are locally peaking and consistent for different modals, and the features with high detected probability are emphasized during the optimization. Different from previous works, the weights are detached from back propagation so that the detected probability of an indistinct feature would not be directly suppressed and the training would be more stable. In this way, our ReDFeat can be readily trained from scratch and adopted in cross-modal image registration. To fulfill the harsh terms of detection in the recoupled constraints and achieve further improvements, we propose the Super Detector that possesses a large receptive field and learnable local non-maximum suppression blocks. Finally, we collect visible and near-infrared, infrared, and synthetic aperture radar image pairs to build a benchmark. Extensive experiments on this benchmark prove the superiority of our ReDFeat and the effectiveness of all proposed components.
\label{sec:sec5}
\IEEEpeerreviewmaketitle

{\small
	\bibliographystyle{plain}
	\bibliography{ReDFeat}
}

\vspace{-0.3in}

\end{document}